\documentclass{article}
\PassOptionsToPackage{numbers}{natbib} 

\usepackage[preprint]{neurips_2026}
\makeatletter
\renewcommand{\@notice}{}
\makeatother

\usepackage[utf8]{inputenc} 
\usepackage[T1]{fontenc}    
\usepackage{url}            
\usepackage{booktabs}       
\usepackage{amsfonts}       
\usepackage{nicefrac}       
\usepackage{microtype}      

\newcommand{\RNum}[1]{\uppercase\expandafter{\romannumeral#1\relax}}
\usepackage{graphicx}
\usepackage{amsfonts}
\usepackage{stix}
\usepackage{multirow}
\usepackage[table,xcdraw]{xcolor}
\usepackage{tabularx}
\usepackage{makecell}
\usepackage{ragged2e}
\usepackage{longtable}
\usepackage{array}
\usepackage{xurl} 
\usepackage{float}
\usepackage{caption}
\usepackage{enumitem}

\usepackage{amsmath}
\usepackage{mathtools}
\usepackage{amsthm}
\usepackage{hyperref} 
\makeatletter

\newcommand{\appendixtoc}{%
  \clearpage
  \phantomsection
  \section*{Appendix}
  \@starttoc{apc}
  \clearpage
}

\let\oldsection\section
\let\oldsubsection\subsection
\let\oldsubsubsection\subsubsection

\newcommand{\appsection}[1]{%
  \oldsection{#1}%
  \addcontentsline{apc}{section}{\protect\numberline{\thesection}#1}%
}

\newcommand{\appsubsection}[1]{%
  \oldsubsection{#1}%
  \addcontentsline{apc}{subsection}{\protect\numberline{\thesubsection}#1}%
}

\newcommand{\appsubsubsection}[1]{%
  \oldsubsubsection{#1}%
  \addcontentsline{apc}{subsubsection}{\protect\numberline{\thesubsubsection}#1}%
}

\makeatother

\title{BrainCoT: A Multi-Task Zero-Shot Brain Signal Foundation Model with Neurometric-Anchored Chain-of-Thought Reasoning}

\author{%
  Yiyu Gui\textsuperscript{1,2},
  Mingzhi Chen\textsuperscript{1,3},
  Guibo Luo\textsuperscript{2}%
  \thanks{Corresponding authors:
  \texttt{luogb@pku.edu.cn, yuchaoyang@pku.edu.cn}.},
  and Yuchao Yang\textsuperscript{1}\footnotemark[1]
  \\[4pt]
  \textsuperscript{1}New Cornerstone Science Laboratory,
  \\
  Guangdong Provincial Key Laboratory of In-Memory Computing Chips,
  \\
  Shenzhen Graduate School, Peking University, Shenzhen, China
  \\[2pt]
  \textsuperscript{2}Guangdong Provincial Key Laboratory of Ultra High Definition
  Immersive Media Technology,
  \\
  Shenzhen Graduate School, Peking University, Shenzhen, China
  \\[2pt]
  \textsuperscript{3}Mohamed bin Zayed University of Artificial Intelligence (MBZUAI), Abu Dhabi, UAE
}

\begin{document}

\maketitle
\raggedbottom 

\begin{abstract}
  Brain signals exhibit substantial multi-center heterogeneity, strong inter-subject variability, and highly diverse task settings, which makes it difficult to build models that generalize beyond a single task. While recent brain signal foundation models leverage large-scale pretraining to learn generalizable representations, they often fall short in practice because they still require target-data-specific supervised adaptation, lack multi-task zero-shot capability, and rarely provide verifiable evidence to support their predictions. In this work, we present \textbf{BrainCoT}, a multi-task zero-shot Brain signal foundation model with neurometric-anchored Chain-of-Thought (CoT) reasoning. BrainCoT unifies heterogeneous tasks into an instruction-driven framework so that, after pretraining, a single model can be directly applied to multiple tasks in zero-shot settings without additional adaptation. It further introduces neurometric-anchored chain-of-thought grounded in neurometric evidence for verifiable reasoning, and adopts Decision-Consistent Rationale Training to keep process learning aligned with correct predictions. Across extensive evaluations, BrainCoT achieves an average ACC of 67.41\% and AUC of 74.33\% in zero-shot classification without downstream training data, outperforming the strongest linear-probing baseline trained with 1\% labeled downstream data by 3.85\% in ACC and 5.97\% in AUC. The code is available at \url{https://github.com/5GYYYYY/BrainCoT}.
\end{abstract}

\section{Introduction}

Brain signals are characterized by multi-center heterogeneity, strong inter-subject variability, and highly diverse task settings, making it difficult for single-task supervised models to generalize. This has recently motivated brain signal foundation models, which aim to learn generalizable representations via large-scale pretraining. However, for a foundation model to be useful in practice for brain signal applications, users typically expect more than ``pretrain then fine-tune'': (i) a single model that can \textbf{be used out of the box across tasks}, enabling zero-shot classification; and (ii) \textbf{reliable explanations} that can support and justify its predictions.

\begin{figure}[tb]
\begin{center}
\centerline{\includegraphics[width=\textwidth]{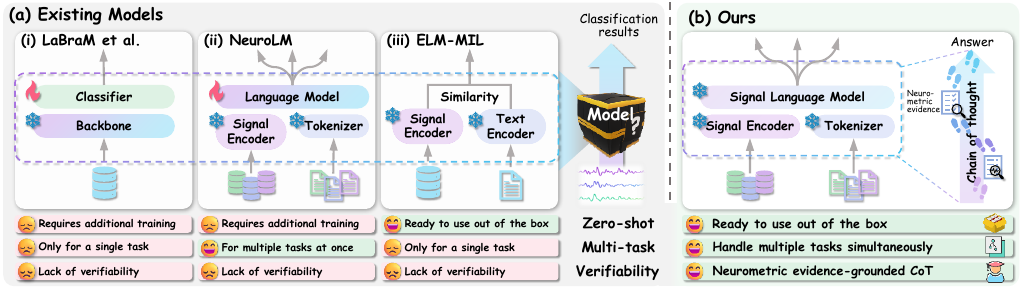}}
\caption{\textbf{Comparison of representative brain signal foundation models and our model.} (a) Existing models fail to achieve multi-task zero-shot generalization and verifiable reasoning, thus lacking both practicality and reliability; (b) Our model enables multi-task zero-shot inference with verifiable reasoning.}
\label{fig1_baseline}
\end{center}
\end{figure}

Despite rapid progress, existing brain signal foundation models still suffer from two major limitations. \textbf{First, practicality is constrained by the lack of multi-task and zero-shot ability.} Most prior models are neither zero-shot nor genuinely multi-task \cite{brainbert,labram,cbramod,csbrain}; instead, they require target-data-specific supervised adaptation, and each adaptation usually targets only one downstream task, as shown in Fig. \ref{fig1_baseline}(a)(i). Some methods support multi-task learning but not zero-shot \cite{neurolm}. Fig.~\ref{fig1_baseline}(a)(ii) illustrates that by introducing language, they can be tuned once over multiple tasks, but this still incurs additional training and therefore does not achieve true out-of-the-box usage. A few recent efforts explore zero-shot inference \cite{elmmil} using a CLIP-style \cite{clip} paradigm for zero-shot classification, but the evaluated scenario remains confined to a single dataset (Fig.~\ref{fig1_baseline}(a)(iii)). \textbf{Second, reliability remains limited: most models output only a classification result}, without an explicit evidence trail that supports the decision. Some attempts provide post-hoc visualizations such as attention heatmaps or t-SNE embeddings, but these cues are often indirect and hard to validate as decision grounds. Recently, Chain-of-Thought \cite{cot} offers a promising path by making the decision process explicit. However, free-form rationales can be non-verifiable and may drift from the actual signal evidence, calling for evidence-anchored reasoning and rationale–answer alignment.

Motivated by these limitations, our goal is to build a brain signal foundation model that is simultaneously (i) \textbf{multi-task and zero-shot}, and (ii) \textbf{capable of evidence-grounded reasoning}, thereby improving both practicality and reliability. However, three major challenges still remain: 

\begin{itemize}[
    leftmargin=1.4em,
    label=\(\bullet\),
    itemsep=0.3em,
    topsep=0.3em,
    parsep=0pt,
    partopsep=0pt
]

\item \textbf{First, to enable multi-task zero-shot usage,} the main bottleneck lies in training resources. Large-scale brain-signal–text corpora covering diverse tasks are still scarce, and reasoning supervision data is even more limited. Moreover, zero-shot capability is not merely a matter of learning strong representations; it requires the model to learn how to map instruction semantics into a decision-making procedure. 

\item \textbf{Second, toward reliable evidence-grounded reasoning,} it is difficult to construct trustworthy reasoning trajectories without detailed clinical reports. Many existing methods rely on LLM-generated chain-of-thought \cite{opentslm,ts_reasoning_1,ts_reasoning_2}, which are hard to verify and prone to hallucinated content that appears plausible but is weakly supported. 

\item \textbf{Third, from the perspective of training paradigm,} achieving both multi-task zero-shot practicality and evidence-grounded reliability requires balancing answer learning and reasoning learning. Strong process supervision can bias the model toward template-like trajectories, whereas weak supervision can diminish trajectory use and reduce the model to answer-only prediction. This trade-off makes it challenging to design a training objective that yields both zero-shot generalization and evidence-grounded reasoning.

\end{itemize}

To address these challenges, we propose \textbf{BrainCoT}, a multi-task zero-shot \textbf{Brain} signal foundation model with neurometric-anchored \textbf{C}hain-\textbf{o}f-\textbf{T}hought reasoning. We make three key contributions: 

\begin{itemize}[
    leftmargin=1.4em,
    label=\(\bullet\),
    itemsep=0.3em,
    topsep=0.3em,
    parsep=0pt,
    partopsep=0pt
]

\item \textbf{BrainCoT enables multi-task zero-shot brain-signal QA.} BrainCoT reformulates heterogeneous brain signal tasks into a unified instruction-driven question answering (QA) framework and pretrains on a large multi-task instruction corpus spanning diverse datasets, enabling out-of-the-box generalization without target-data-specific supervised adaptation. 

\item \textbf{BrainCoT supports reliable inference with verifiable reasoning.} Instead of using unverifiable LLM-generated rationales, BrainCoT constructs neurometric-anchored CoT from neurometric evidence extracted from the input signals, making the reasoning trajectory externally checkable. 

\item \textbf{BrainCoT aligns process learning with answer quality.} BrainCoT introduces decision-consistent rationale tuning that couples process learning with answer quality, encouraging trajectories that help produce correct answers while reducing template imitation and trajectory–answer mismatch.

\end{itemize}


\begin{figure*}[htb]
\begin{center}
\centerline{\includegraphics[width=\textwidth]{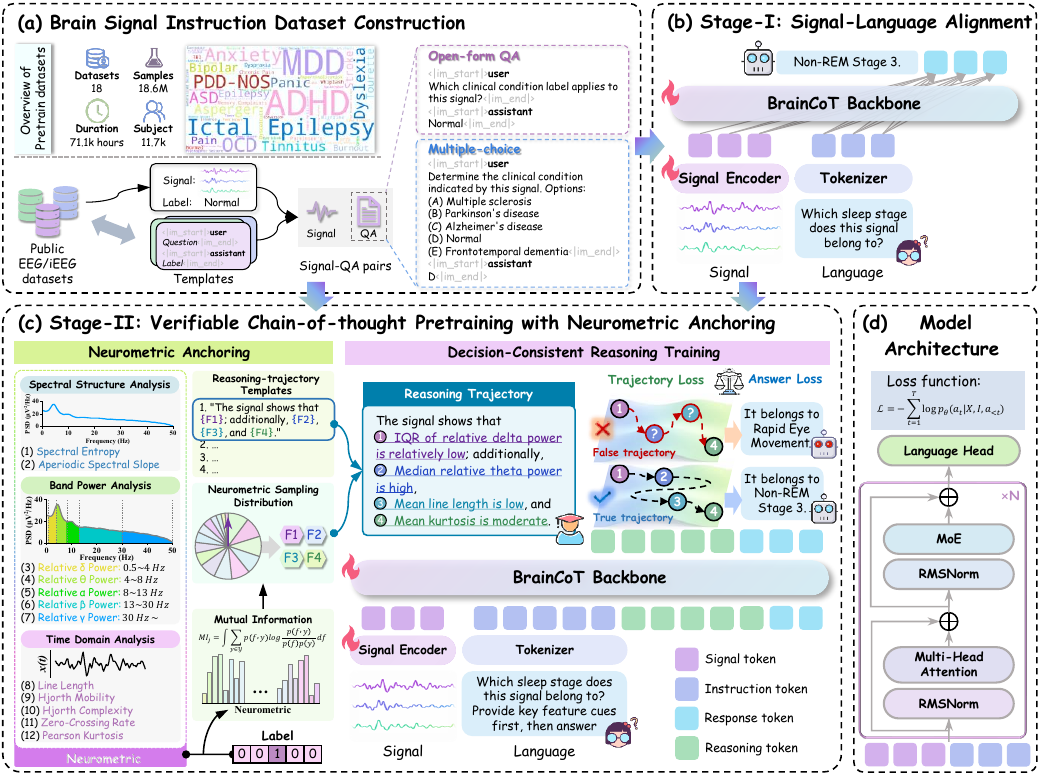}}
\caption{\textbf{Overview of BrainCoT.} (a) Dataset construction. We collect 18.6M signal segments from 18 EEG/iEEG datasets spanning diverse tasks, and convert them into instruction-following signal–QA pairs using task-specific templates for pretraining. (b) Stage-\RNum{1}: Signal-Language alignment. BrainCoT is pretrained to answer task queries from signal inputs under a unified signal–language QA formulation, establishing multi-task instruction following and direct answering capability. (c) Stage-\RNum{2}: Verifiable chain-of-thought pretraining with neurometric-anchoring. We generate neurometric-anchored reasoning trajectories from neurometric evidence and perform decision-consistent reasoning training, enabling verifiable chain-of-thought inference. (d) The model architecture of BrainCoT.}
\label{fig2_ourmodel}
\end{center}
\end{figure*}

\vspace{-0.4cm}

\section{Method}
Our goal is to build BrainCoT, a multi-task zero-shot brain signal foundation model that generalizes across diverse tasks without additional target-data-specific adaptation, while producing verifiable reasoning trajectories. We cast diverse brain-signal tasks into a unified instruction-driven Brain Signal QA formulation: given a signal segment $X$ and an instruction $I$, the model autoregressively generates a textual answer $A$, using language as a shared semantic interface to unify heterogeneous tasks and datasets. Our approach follows a three-part pipeline.

\textbf{(1) Dataset Construction.}
We convert labeled brain-signal datasets into paired $(X,I,A)$ training triples using task-specific templates, yielding a large-scale instruction corpus.

\textbf{(2) Stage-\RNum{1} pretraining.} We pretrain a Brain Signal QA model to generate $A$ conditioned on $(X,I)$; the resulting model enables \textbf{Direct Mode} for efficient inference.

\textbf{(3) Stage-\RNum{2} pretraining.}
We further train the model with supervision from verifiable reasoning trajectories $R$ constructed by neurometric anchoring, so that it can first generate trajectories $R$ and then answer $A$. This yields the \textbf{Reasoning Mode}, improving accuracy and providing verifiable reasoning trajectories that support the model’s predictions.

\subsection{Brain Signal Instruction Dataset Construction}

Building an instruction-following Brain Signal QA model requires paired signal-instruction-answer data, yet such corpora are scarce in the brain signal domain. We therefore construct a scalable signal–language QA pretraining set via a template-based pipeline.

Let the $i$-th dataset be $\mathcal{D}^{(i)}={{(x_n^{(i)},y_n^{(i)})}}_{n=1}^{N_i}$. For each dataset, we design a task-specific template set $\mathcal{T}^{(i)}$. Given a sample, we randomly draw a template $\mathscr{t}\sim\mathcal{T}^{(i)}$ to create an instruction $I=\mathscr{t}_I(\cdot)$ and a textual answer $A=\mathscr{t}_A(y)$, forming a training triple $(X,I,A)$. Here, $y$ denotes the task label, and the answer $A$ is produced by applying label-to-text rules defined by the selected template. The full template specification is provided in the Appendix \ref{app_qa_template}. Our templates support both multiple-choice and open-form QA under a unified autoregressive objective. This pipeline converts 18 datasets into over 18.6M brain-signal-QA pairs for Stage-\RNum{1} pretraining.

\subsection{Stage-\RNum{1}: Signal-Language Alignment}

As illustrated in Fig.~\ref{fig2_ourmodel}(b), Stage-\RNum{1} aligns brain signals with language by training a base brain-signal QA model. Given a brain signal segment $X\in\mathbb{R}^{B\times C\times L}$, where $B$ is the batch size, $C$ is the number of channels, and $L$ is the number of sampled time points, respectively, the model receives the corresponding textual instruction $I$ (e.g., “Which sleep stage does this segment belong to?”). Conditioned on $(X,I)$, the model autoregressively generates an answer sequence $A=(a_1,a_2,...,a_T)$, learning the conditional distribution:
\vspace{-0.1cm}
\begin{equation}
p_\theta(A|X,I)=\prod_{t=1}^{T}{p_\theta(a_t|X,I,a_{<t})}
\end{equation}
\vspace{-0.2cm}
During training, we optimize cross-entropy only over the answer tokens:
\begin{equation}
\mathcal{L}_{ans}=-\sum_{t=1}^{T}{logp_\theta(a_t|X,I,a_{<t})}
\end{equation}
After Stage-\RNum{1} pretraining, the model supports brain signal QA and can be directly applied for \textbf{Direct Mode} inference.

\subsection{Stage-\RNum{2}: Verifiable Chain-of-Thought Pretraining with Neurometric-Anchoring}

\subsubsection{Verifiable Reasoning-trajectory Construction via Neurometric Anchoring}

As discussed in Appendix~\ref{app:related_work}, prior time-series CoT methods often rely on free-form LLM rationales. In contrast, we emphasize verifiability for brain signal reasoning: the reasoning trajectory should be grounded in deterministic, reproducible analyses of the input rather than unverifiable narratives. We aim to make the model behave more like neuroscience experts: extract interpretable neurometric evidence first, then decide from it.

To this end, we propose a neurometric anchoring scheme to construct verifiable reasoning trajectories and to reduce reasoning drift by selecting label-relevant evidence. Specifically, we compute a fixed set of neurometric features commonly used in neuroscience (as shown in the left panel of Fig.~\ref{fig2_ourmodel}(c)): (i) Spectral Structure Analysis: \textit{Spectral Entropy} and \textit{Aperiodic Spectral Slope}; (ii) Band Power Analysis: \textit{Relative $\delta$}, $\theta$, $\alpha$, $\beta$, and \textit{$\gamma$ Power}; (iii) Time Domain Analysis: \textit{Line Length}, \textit{Hjorth Mobility}, \textit{Hjorth Complexity}, \textit{Zero-Crossing Rate}, and \textit{Pearson Kurtosis}. Detailed definitions and implementations are provided in Appendix \ref{app_neurometric}.

We consider a set of representative statistical feature functions $\{{f_j(\cdot)}\}_{j=1}^M$, and compute candidate evidence from the signal $F_j=f_j(X)$. To suppress irrelevant features, we compute the mutual information between each neurometric feature and the label:
\vspace{-0.1cm}
\begin{equation}
{MI}_j=\int\sum_{y\in\mathcal{Y}}{p(f,y)log\frac{p(f,y)}{p(f)p(y)}df}
\end{equation}
We estimate ${MI}_j$ from each dataset’s empirical distribution and use it as a relevance score to define the evidence-sampling distribution (see Appendix~\ref{MI_stimation} for implementation details). We then map the scores to a sampling distribution:
\vspace{-0.2cm}
\begin{equation}
\mathscr{p}_j=\frac{exp(\alpha{MI}_j)}{\sum_{m=1}^{M}{{exp(\alpha MI}_m)}}
\end{equation}
where temperature parameter $\alpha$ controls the sharpness of $\mathscr{p}$. We sample $K$ distinct feature indices according to $p$, forming a set $\mathcal{S}\subset\{1,2,...,M\}$, and use $\{{F_j}\}_{j\in\mathcal{S}}$ as evidence elements in the trajectory.

Instead of using raw feature values, we discretize each feature into quantile-based rank levels, improving robustness across subjects and centers and matching how neuroscience experts describe evidence (e.g., low/high). Let $\{{q_j^{(u)}}\}_{u=1}^U$ denote the quantile thresholds for feature $j$. We define its level as
\vspace{-0.2cm}
\begin{equation}
\ell_j=bin(F_j;\{{q_j^{(u)}}\}_{u=1}^U)
\end{equation}
As shown in Fig.\ref{fig2_ourmodel}(c), we fill a reasoning-trajectory template with the selected evidence $\{{(j,\ell_j)}\}_{j\in\mathcal{S}}$ to generate the textual reasoning trace $R$. Since each evidence statement in $R$ is deterministically computed from $X$, the trajectory is inherently verifiable and provides reliable supervision for Stage-\RNum{2}. This neurometric anchoring pipeline is used only to construct offline supervision for Stage-\RNum{2}; at inference time, BrainCoT does not receive externally computed neurometric features, MI scores, or quantile levels, but generates the reasoning trajectory directly from the signal tokens and instruction tokens. Full template details are in Appendix~\ref{app_reason_trace_template}.

\subsubsection{Decision-Consistent Reasoning Training}

Stage-\RNum{2} fine-tunes the Stage-\RNum{1} base model to support reasoning using verifiable trajectories as supervision. During training, the model is optimized to generate both the reasoning trajectory and the final answer. A naive combination of the trajectory loss and the answer loss with a fixed weight can lead to a length bias: samples with longer trajectories produce larger gradients, which encourages superficial trajectory imitation and weakens the decision linkage from trajectory to answer.

As shown in Fig.~\ref{fig2_ourmodel}(c) (right), we address this issue with a decision-consistent training mechanism. The key idea is to adjust how strongly the model learns the trajectory according to its current answer quality, so that trajectory learning focuses on evidence patterns that actually help produce correct answers. Let $\mathcal{T}_R$ denote token positions of the evidence trace and $\mathcal{T}_A$ denote token positions of the answer. We define
\begin{equation}
\begin{split}
\mathcal{L}_{ans}=-\sum_{t\in\mathcal{T}_A}{logp_\theta(y_t|X,I,y_{<t})},  \mathcal{L}_{think}=-\sum_{t\in\mathcal{T}_R}{logp_\theta(y_t|X,I,y_{<t})}
\end{split}
\end{equation}
We construct a reasoning weight from the answer loss:
\begin{equation}
\omega=exp(\frac{-sg(\mathcal{L}_{ans})}{\tau})
\end{equation}
where $sg(\cdot)$ denotes stop-gradient and $\tau>0$ controls training stability in early epochs. The final objective is
\begin{equation}
\mathcal{L}_{total}=\mathcal{L}_{ans}+\omega\mathcal{L}_{think}
\end{equation}
With this design, trajectory learning strength adapts to decision quality, thereby suppressing purely formal imitation and strengthening the decision relevance between the reasoning trajectory and the final prediction.

\subsection{Model Architecture}
As shown in Fig.~\ref{fig2_ourmodel}(d), BrainCoT uses a decoder-only Transformer architecture for instruction-driven brain-signal analysis. A universal signal encoder maps heterogeneous brain signals into signal tokens, which are concatenated with instruction tokens and fed into a shared backbone for autoregressive modeling. We further adopt a Mixture-of-Experts (MoE) backbone to enable conditional computation under diverse tasks. Additional details are provided in Appendix~\ref{app_model_arch}.

\subsection{Brain Signal Classification via Candidate-Token Scoring}

We reformulate brain signal classification as a multiple-choice question answering problem, where prediction is performed by scoring candidate answer tokens under next-token prediction. Let $\mathcal{V}$ denote the vocabulary. For a $K$-class task, we define a candidate answer token set $\mathcal{C}={c_1,c_2,...,c_K}\subset\mathcal{V}$. Given the input $(X,I)$, the model outputs a logit vector $z\in\mathbb{R}^{\left|\mathcal{V}\right|}$ at the answer-start position. We compute probabilities over the candidate set as the classification distribution:
\begin{equation}
P({y=c}_i|X)=\frac{exp(z_{c_i})}{\sum_{j=1}^{K}{exp(z_{c_j})}},\ \ i=1,2,...,K
\end{equation}
The prediction is then given by the highest-probability candidate token:
\begin{equation}
\hat{y}=argmaxP(y=c_k|X,I)
\end{equation}

\section{Experimental Setup}

\subsection{Pretraining}
\textbf{Data preprocessing.}
We apply an identical preprocessing pipeline for both pretraining and downstream evaluation. Specifically, all signals are resampled to 200\,Hz and filtered with a 50/60 Hz notch filter (depending on the acquisition region) to suppress power-line interference. 

\textbf{Pretraining datasets.}
We curate a large-scale pretraining corpus covering two brain signal modalities, including scalp EEG and intracranial EEG (iEEG). Overall, it contains \textbf{18 datasets}, including \textbf{14} EEG datasets \cite{shhs,seizeit2,tusz,tuab,chbmit,tuep,mdd,hmc,tdbrain,tuev,adftd,brainlat,ad-auditory,reeg-pd} and \textbf{4} iEEG datasets \cite{hup,upenn,swecethz,mayo}. In total, it provides over \textbf{70{,}000 hours} of recordings from more than \textbf{11{,}000 subjects}, spanning diverse acquisition settings (see Appendix \ref{app_pretraining_datasets} for detailed dataset statistics). 

\subsection{Evaluation}

\textbf{Downstream datasets.}
As summarized in Table~\ref{tab:downstream_datasets}, we evaluate BrainCoT on \textbf{12} downstream datasets spanning iEEG and EEG modalities. More details are provided in the Appendix \ref{app_downstream_datasets}.

\begin{table*}[htb]
\caption{\textbf{Overview of downstream datasets.} ``Label'' denotes the number of classes; ``Channels'', ``Subjects'', and ``Duration'' denote channel count, subject count, and per-sample length, respectively.}
\label{tab:downstream_datasets}
\centering
\small
\setlength{\tabcolsep}{6pt}
\renewcommand{\arraystretch}{1.08}
\resizebox{\textwidth}{!}{%
\begin{tabular}{@{}c c c c c c c c@{}}
\toprule
\textbf{Signal modality} & \textbf{Dataset} & \textbf{Task} & \textbf{Label} & \textbf{Channels} & \textbf{Duration} & \textbf{Subjects} & \textbf{Origin sample rate} \\
\midrule
\multirow{2}{*}{iEEG}
& Mayo \cite{mayo}           & Interictal epileptiform activity detection (IEA detection) & 2-class & 1  & 3s  & 15 & 500 Hz \\
& IEDS \cite{ieds}           & Interictal epileptiform activity detection (IEA detection) & 2-class & 1  & 3s  & 25 & 1000 Hz \\
\midrule
\multirow{10}{*}{EEG}
& ISRUC \cite{isruc}         & Sleep stage classification        & 5-class & 2  & 30s & 100 & 200 Hz \\
& Dreams \cite{dreams}       & Sleep stage classification        & 5-class & 3  & 30s & 20  & 200 Hz \\
& Siena \cite{siena}         & Epileptic state recognition       & 4-class & 16 & 10s & 14  & 512 Hz \\
& SanDiego \cite{sandiego}   & Parkinson's disease detection     & 2-class & 32 & 2s  & 31  & 512 Hz \\
& ADHD \cite{adhd-121}       & Attention deficit hyperactivity disorder (ADHD) detection & 2-class & 19 & 5s  & 121 & 128 Hz \\
& Schizo \cite{schizo-youth} & Schizophrenia detection           & 2-class & 16 & 5s  & 84  & 128 Hz \\
& APAVA \cite{apava}         & Alzheimer's disease detection     & 2-class & 16 & 2s  & 23  & 256 Hz \\
& ADFSU \cite{adfsu}         & Alzheimer's disease detection     & 2-class & 19 & 2s  & 92  & 128 Hz \\
& NMT \cite{nmtab}           & Anomaly detection                 & 2-class & 16 & 10s & 2417  & 200 Hz \\
& SEE \cite{see}             & Clinical diagnosis classification (epilepsy)                 & 2-class & 1  & 4s  & 30  & 173.61 Hz \\
\bottomrule
\end{tabular}%
}
\end{table*}

\textbf{Baselines.}
We compare BrainCoT with representative brain signal foundation models, including BrainBERT \cite{brainbert}, BIOT \cite{biot}, LaBraM \cite{labram}, CBraMod \cite{cbramod}, NeuroLM \cite{neurolm}, ELM-MIL \cite{elmmil}, and CSBrain \cite{csbrain}. For fair comparison, BrainCoT and all baselines use the same preprocessing and evaluation strategy. Details of these baselines can be found in Appendix \ref{app_baseline}.

\textbf{Downstream settings.}
We use subject-wise 5-fold cross-validation for all datasets, except NMT \cite{nmtab} and SEE \cite{see}. In each run, 3 folds are used for training, one for validation, and one for testing; all segments from the same subject are kept in a single fold to prevent subject leakage. NMT and SEE provide predefined train/test splits; we further split the original training set into train/val with an 8:2 ratio. We report test-set ACC and macro AUC averaged over five folds.

\textbf{Zero-shot protocol and baseline reference.} 
In this work, zero-shot evaluation refers to a no-downstream-adaptation protocol: after pretraining, BrainCoT is directly applied to each downstream dataset without parameter updates or task-specific use of downstream training labels. We formulate downstream classification as signal-language QA, where the model takes a signal segment and textual instruction as input and generates the answer. Since existing baselines cannot be directly evaluated under the same native zero-shot protocol, we use supervised low-label settings as reference points. Specifically, ELM-MIL~\cite{elmmil} claims zero-shot capability but does not release the pretrained text-side weights required for zero-shot inference, while the other baselines do not support native zero-shot prediction. Therefore, following prior zero-shot studies~\cite{clip,zeroshot_compare}, we compare BrainCoT zero-shot results with baseline performance under 1\% and 10\% linear probing as supervised reference points.
 
\vspace{-0.15cm}

\subsection{Implementation Details}

We pretrain the model in \textbf{two stages}. In each stage, we use a per-dataset batch size of 128, resulting in a total batch size of 2304, and train for 1 epoch. The peak learning rate is set to $2\times10^{-4}$ for \textbf{Stage-\RNum{1}} and $2\times10^{-5}$ for \textbf{Stage-\RNum{2}}. We adopt a cosine learning-rate schedule with a 10\% iteration warm-up. Pretraining is performed in bfloat16 precision for computational efficiency. For downstream experiments, we train with batch size 128, learning rate $2\times10^{-4}$, and float32 precision, using early stopping on validation AUC (patience $=10$, up to 100 epochs). All experiments were conducted on a Linux server equipped with an NVIDIA A100 80GB GPU, using PyTorch 2.0.1 and CUDA 12.2. 

\section{Results}

\subsection{Zero-shot Performance}

We evaluate BrainCoT in a zero-shot setting across 12 downstream EEG/iEEG datasets. 
Table \ref{ours_zero_shot} reports the absolute performance on each dataset, demonstrating that BrainCoT achieves strong results without any target-data-specific adaptation, with an average ACC of 67.41\% and AUC of 74.33\%.

\begin{table}[htb]
\caption{\textbf{Zero-shot performance across datasets.} Average ACC = 67.41, Average AUC = 74.33.}
\label{ours_zero_shot}
\resizebox{\textwidth}{!}{%
\setlength{\tabcolsep}{3pt}
\renewcommand{\arraystretch}{1.25}
\begin{tabular}{@{}ccccccccccccc@{}}
\toprule
\textbf{Dataset} & \textbf{ISRUC} & \textbf{Dreams} & \textbf{Siena} & \textbf{SanDiego} & \textbf{ADHD} & \textbf{Schizo} & \textbf{APAVA}      & \textbf{ADFSU}       & \textbf{IEDS}       & \textbf{Mayo}       & \textbf{NMT}        & \textbf{SEE}        \\ \midrule
ACC       & 82.77$_{\pm 1.38}$     & 77.66$_{\pm 2.51}$      & 47.31$_{\pm 16.20}$    & 62.64$_{\pm 9.65}$        & 59.13$_{\pm 3.37}$    &   62.23$_{\pm 6.49}$    & 73.20$_{\pm 12.71}$ & 80.05$_{\pm 8.67}$ & 67.80$_{\pm 6.32}$ & 77.26$_{\pm 6.76}$ & 57.47$_{\pm 0.00}$ & 61.45$_{\pm 0.00}$ \\
AUC       & 95.19$_{\pm 0.62}$     & 92.49$_{\pm 1.07}$      & 64.85$_{\pm 9.68}$     & 63.61$_{\pm 11.13}$       & 59.44$_{\pm 3.83}$    & 73.53$_{\pm 8.35}$      & 82.36$_{\pm 6.60}$  & 71.79$_{\pm 15.45}$ & 61.00$_{\pm 4.98}$ & 88.31$_{\pm 4.24}$ & 75.10$_{\pm 0.00}$ & 64.24$_{\pm 0.00}$ \\ \bottomrule
\end{tabular}%
}
\end{table}

To further assess its relative performance, Table \ref{zero_shot_comparison} summarizes the comparison with baseline models under 1\% and 10\% linear probing. \underline{Underline} denotes the strongest non-BrainCoT baseline. BrainCoT outperforms the strongest 1\% linear probing baseline by 3.85\% in ACC and 5.97\% in AUC, while remaining competitive with 10\% linear probing. This indicates that its pretrained representations already capture substantial task-relevant information. Per-dataset results are provided in Appendix \ref{detail_zero-shot}.

\begin{table}[!ht]
\caption{\textbf{Comparison with linear probing baselines (average performance).}}
\label{zero_shot_comparison}
\resizebox{\textwidth}{!}{%
\setlength{\tabcolsep}{6pt}
\begin{tabular}{@{}cccccccccc@{}}
\toprule\rowcolor{gray!15}
\multicolumn{2}{c}{\textbf{Model}} & \textbf{BrainBERT}   & \textbf{BIOT}        & \textbf{LaBraM }    & \textbf{CBraMod}     & \textbf{NeuroLM}     & \textbf{ELM-MIL}     & \textbf{CSBrain}    & \textbf{BrainCoT}     \\ 
Metric & Training data      & ICLR' 23           & NeurIPS' 23           & ICLR' 24           & ICLR' 25           & ICLR' 25           & ICML' 25           & NeurIPS' 25          & ours                                    \\ \midrule&  0\% (zero-shot)                    &      -          &     -        &    -         &      -       &     -        &     -       &       -    & \textbf{67.41}                                          \\
                                     \multirow{-1}{*}{ACC $\uparrow$}& 1\%                    & 62.47    & \underline{63.56}    & 59.24    & 57.58    & 57.43    & 59.56    & 55.06   & \textbf{70.74}    \\
                &  10\%                   & \underline{67.69}    & 67.15    & 63.15     & 64.61    & 62.52     & 63.27    & 56.00   & \textbf{73.78}    \\ \cmidrule(lr){2-10}
&  0\% (zero-shot)                    &      -          &     -        &    -         &      -       &     -        &     -       &       -    & \textbf{74.33}                                        \\
                                     \multirow{-1}{*}{AUC $\uparrow$}& 1\%                    & \underline{68.36}    & 63.02    & 57.54    & 59.43    & 58.15    & 61.36    & 52.92   & \textbf{75.59}   \\
&  10\%                   & \underline{73.23}    & 71.59    & 62.09     & 68.36    & 65.43     & 66.75    & 54.19   & \textbf{80.19}    \\ \bottomrule
\end{tabular}%
}
\end{table}

\begin{figure*}[htb]
\vspace{-0.1cm}
\begin{center}
\centerline{\includegraphics[width=\textwidth]{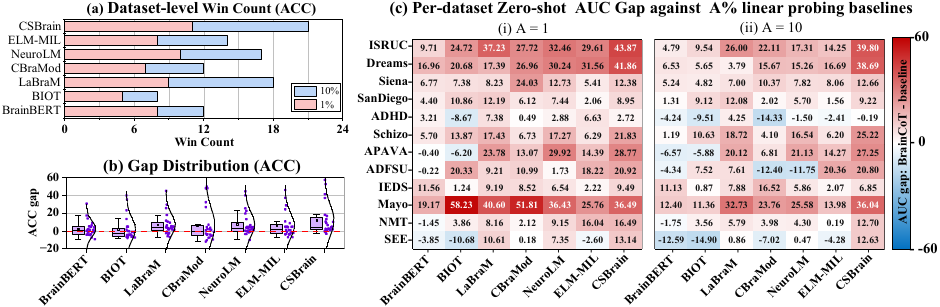}}
\caption{\textbf{Multi-level comparison between BrainCoT zero-shot inference and linear probing baselines.} \textbf{(a) ACC win count}: the number of datasets where BrainCoT zero-shot outperforms each baseline under 1\% and 10\% linear probing. \textbf{(b) ACC gap distribution}: the distribution of performance gaps between BrainCoT zero-shot and each baseline across datasets, where positive values indicate BrainCoT performs better. \textbf{(c) AUC gap heatmaps}: dataset-level AUC differences between BrainCoT zero-shot and each baseline under 1\% linear probing (i) and 10\% linear probing (ii); \textcolor{red}{red} denotes positive gaps and \textcolor{blue}{blue} denotes negative gaps.}
\label{zero_shot_heat}
\end{center}
\end{figure*}

Fig.~\ref{zero_shot_heat} further complements Table~\ref{zero_shot_comparison} by showing that the advantage of BrainCoT is not only reflected in the averaged results, but also appears consistently across baselines and datasets. In Fig.~\ref{zero_shot_heat}(a), BrainCoT shows high ACC win counts against most baselines, especially under the 1\% linear probing setting. Under 10\% linear probing, the win counts decrease but remain substantial for several baselines, suggesting that BrainCoT is still competitive against stronger supervised references. In Fig.~\ref{zero_shot_heat}(b), the ACC gap distributions are mostly centered above zero, showing that the advantage is not driven by a few datasets. Positive medians are clear for LaBraM, NeuroLM, ELM-MIL, and CSBrain, while smaller margins for BrainBERT and BIOT suggest variation across architectures. In Fig.~\ref{zero_shot_heat}(c), the AUC gap heatmaps provide a dataset-level view. Under 1\% linear probing, most dataset-baseline pairs show positive gaps, confirming the broad zero-shot advantage of BrainCoT. Under the stronger 10\% setting, BrainCoT remains competitive on datasets such as ISRUC, Dreams, Mayo, and IEDS. The few negative gaps indicate that 10\% supervised adaptation can still be beneficial in certain cases, but the overall trend still supports BrainCoT's strong zero-shot transfer across EEG/iEEG tasks.

\subsection{Effect of Reasoning}

To validate chain-of-thought (CoT) in brain signal analysis, we conduct quantitative and qualitative analyses. As summarized in Table~\ref{result_reasoning_analysis}, BrainCoT with Stage-\RNum{2} neurometric-anchored reasoning consistently improves over the same Stage-\RNum{2} model in Direct mode, increasing ACC and AUC by 4.50\% and 4.77\%, respectively. Per-dataset results are provided in Table~\ref{app_result_reasoning_analysis} in Appendix. Stage-\RNum{2} training is also meaningful: even without reasoning at test time, Stage-\RNum{2} Direct substantially outperforms Stage-\RNum{1} Direct, suggesting that second-stage reasoning pretraining enhances the backbone’s task understanding rather than simply encouraging longer outputs. Finally, the reasoning intervention, which disrupts the reasoning trajectory by randomly replacing neurometric names and intensities, markedly degrades performance, with drops of 6.70\% in ACC and 6.37\% in AUC relative to Stage-\RNum{2} Reasoning. This suggests that reasoning is not merely a formality but provides decision-relevant evidence that influences model predictions.

\begin{table}[htb]
\caption{\textbf{The average performance under different reasoning settings.}}
\label{result_reasoning_analysis}
\setlength{\tabcolsep}{4pt}
\resizebox{\textwidth}{!}{%
\begin{tabular}{@{}ccccccc@{}}
\toprule
\textbf{Variants}    & \textbf{Stage-\RNum{1}} & \textbf{Stage-\RNum{2}} & \textbf{Reasoning} & \textbf{Correct reasoning trajectory} & \textbf{ACC $\uparrow$} & \textbf{AUC $\uparrow$} \\ \midrule
Stage-\RNum{1} Direct       & $\checkmark$                &                  &                    &                                       & 57.59$_{\textcolor{green}{\downarrow\,9.82}}$        & 63.18$_{\textcolor{green}{\downarrow\,11.15}}$        \\
Stage-\RNum{2} Direct       & $\checkmark$                & $\checkmark$                &                    &                                       & 62.91$_{\textcolor{green}{\downarrow\,4.50}}$        & 69.56$_{\textcolor{green}{\downarrow\,4.77}}$        \\
Stage-\RNum{2} Intervention & $\checkmark$               & $\checkmark$                & $\checkmark$                  &                                       & 60.71$_{\textcolor{green}{\downarrow\,6.70}}$        & 67.96$_{\textcolor{green}{\downarrow\,6.37}}$        \\
\rowcolor{gray!15} Stage-\RNum{2} Reasoning (full BrainCoT)   & \checkmark                & $\checkmark$                & $\checkmark$                  & $\checkmark$                                     & \textbf{67.41}        & \textbf{74.33}       \\ \bottomrule
\end{tabular}%
}
\end{table}

We further validate the informativeness of the extracted neurometric features by training conventional machine-learning classifiers on these features. As shown in Table~\ref{tab:ml_baselines} of Appendix~\ref{app_reasoning_analysis}, these feature-only classifiers achieve clearly above-chance performance, indicating that the neurometric features contain discriminative information for downstream classification. This supports their use as meaningful reasoning evidence rather than merely template-based textual cues.

\begin{figure*}[htb]
\begin{center}
\centerline{\includegraphics[width=\textwidth]{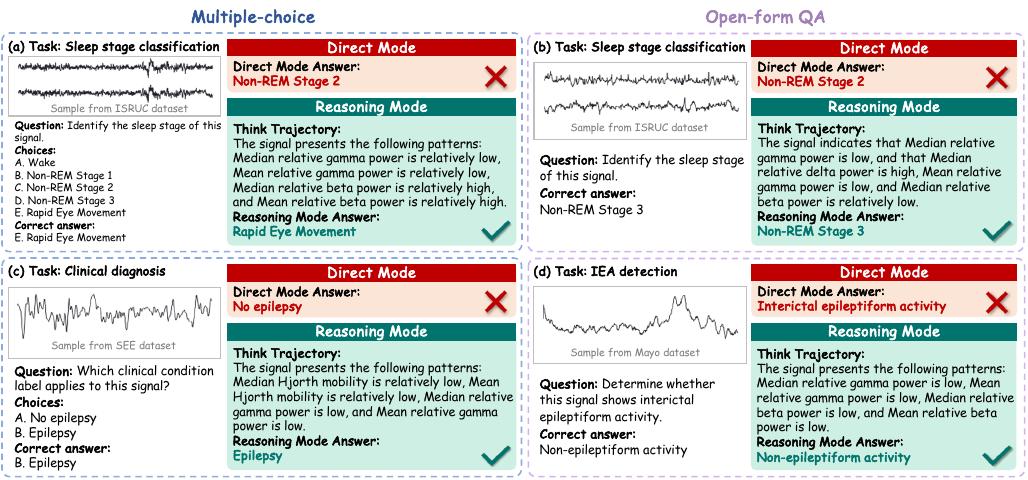}}
\caption{\textbf{Examples of Direct vs. Reasoning Inference.} Representative cases on sleep stage classification and epilepsy-related tasks, comparing predictions from Direct and Reasoning modes.}
\label{fig3_reasoning}
\end{center}
\end{figure*}

Following the quantitative ablations, we provide qualitative evidence in Fig.~\ref{fig3_reasoning}. Each panel shows the same input sample evaluated under two inference modes: Direct Mode produces an incorrect prediction, whereas Reasoning Mode generates a neurometric-grounded reasoning trajectory and returns the correct answer. Across tasks and formats (multiple-choice and open-form QA), we observe that the reasoning trajectories are consistent with signal features. For example, in Fig.~\ref{fig3_reasoning}(a), Direct Mode misclassifies the segment as Non-REM Stage 2. In contrast, Reasoning Mode explicitly identifies lower gamma power and higher beta power, and accordingly predicts Rapid Eye Movement (REM). This is physiologically plausible, since elevated beta activity is one of the characteristic cues that human experts often consider during sleep staging. These cases suggest that BrainCoT does not merely “talk through” a decision, but can extract discriminative frequency-band evidence and use it to correct direct-decoding errors, yielding predictions that are both more accurate and interpretable.

\subsection{Full Fine-tuning and Linear Probing Performance}

Table \ref{more_result} summarizes the average performance of BrainCoT and baselines across downstream datasets under full fine-tuning and linear probing. BrainCoT achieves the best average ACC and AUC under linear probing. Under full fine-tuning, BrainCoT also shows competitive performance, achieving the best ACC and an AUC second only to ELM-MIL but higher than the remaining baselines. These results suggest that BrainCoT performs well in performance-oriented full fine-tuning with a stronger advantage under lightweight linear probing. Per-dataset results are reported in Appendix \ref{app_linear_probing} and \ref{app_full_finetuning}.

\begin{table}[htb]
\caption{\textbf{Average full fine-tuning and linear probing performance across 12 datasets.} \textbf{Bold} denotes the best performance and \underline{underline} denotes the second best performance. }
\label{more_result}
\resizebox{\textwidth}{!}{%
\begin{tabular}{@{}cccccccccc@{}}
\toprule
\multicolumn{1}{l}{}            & \textbf{Metric} & \textbf{BrainBERT} & \textbf{BIOT}  & \textbf{LaBraM} & \textbf{CBraMod} & \textbf{NeuroLM}        & \textbf{ELM-MIL}        & \textbf{CSBrain}        & \textbf{~~~BrainCoT~~~~~}       \\ \midrule
\multirow{2}{*}{Full fine-tuning}  & ACC $\uparrow$   & 68.98     & 75.28 & 73.43   & \underline{75.78}   & 68.06          & 74.64          & 71.15 & $\textbf{77.40}_{\textcolor{red}{\uparrow\,1.62}}$ \\
                                & AUC $\uparrow$   & 76.46     & 84.12 & 79.11   & 82.82   & 74.34 & \textbf{85.95} & 74.52          & $\underline{85.01}_{\textcolor{green}{\downarrow\,0.94}}$          \\ \midrule
\multirow{2}{*}{Linear probing} & ACC $\uparrow$   & 68.55     & 65.44 & 64.05   & \underline{70.51}   & 60.39          & 64.36          & 67.37          & $\textbf{75.42}_{\textcolor{red}{\uparrow\,4.91}}$ \\
                                & AUC $\uparrow$   & \underline{76.55}     & 75.33 & 65.01   & 72.82   & 71.00          & 70.62          & 57.33          & $\textbf{82.62}_{\textcolor{red}{\uparrow\,6.07}}$ \\ \bottomrule
\end{tabular}%
}
\end{table}

\subsection{Scaling Model Size}

\begin{figure}[H]
\centering

\begin{minipage}[t]{0.43\textwidth}
\vspace{0pt}
\centering
\small
\captionof{table}{\textbf{Model configurations of BrainCoT at different scales.}}
\label{tab:model_size}
\vspace{-0.2cm}
\renewcommand{\arraystretch}{1}
\resizebox{\linewidth}{!}{%
\begin{tabular}{lccc}
\toprule
\textbf{Model} & \textbf{Small} & \textbf{Medium} & \textbf{Base} \\
\midrule
Hidden dimension & 384 & 512 & 768 \\
FFN dimension & 1536 & 2048 & 3072 \\
Transformer layers & 6 & 8 & 12 \\
Attention heads & 6 & 8 & 12 \\
\rowcolor{gray!15} Total parameters & 83.8M & 139.2M & 329.2M \\
\bottomrule
\end{tabular}
}
\end{minipage}
\hfill
\begin{minipage}[t]{0.54\textwidth}
\vspace{0pt}
\centering
\begingroup
\setbox0=\hbox{\includegraphics[width=\linewidth]{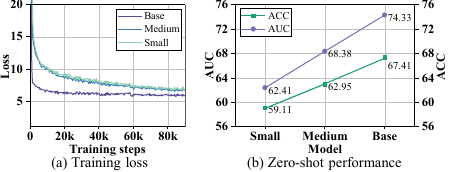}}%
\immediate\pdfxform0
\leavevmode\pdfrefxform\pdflastxform
\endgroup\par
\captionof{figure}{\textbf{Scaling Analysis.}}
\label{fig:scaling}
\end{minipage}

\end{figure}

To examine how model size affects BrainCoT, we conduct a scaling experiment. Table~\ref{tab:model_size} lists the key configurations and parameter counts of the three BrainCoT variants, and Fig.~\ref{fig:scaling} shows their zero-shot scaling performance. As the model size increases from BrainCoT-Small (83.8M) to BrainCoT-Medium (139.2M) and BrainCoT-Base (329.2M), training loss decreases more stably, while zero-shot ACC and AUC show consistent improvements. These results indicate that larger BrainCoT models have stronger pretraining capacity and better generalization, supporting further scaling of model size.

\clearpage
\subsection{Ablation Study}

\begin{figure}[H]
\begin{center}
\centerline{\includegraphics[width=\columnwidth]{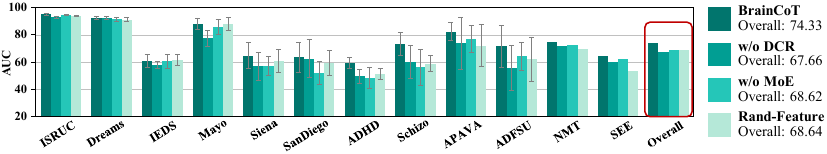}}
\caption{\textbf{Ablation study on 12 downstream datasets.}}
\label{fig6_ablation}
\end{center}
\end{figure}

We compare zero-shot BrainCoT with three ablations: w/o DCR (Decision-Consistent Reasoning), w/o MoE, and Rand-Feature (random evidence instead of MI-weighted sampling). Across 12 datasets, the full model outperforms all variants (Fig.~\ref{fig6_ablation}), indicating that each component contributes to zero-shot generalization. Removing DCR leads to the largest drop, highlighting its key role in aligning reasoning with correct decisions. Dropping MoE also degrades performance, suggesting the importance of expert routing. Replacing MI-weighted evidence sampling with random features further reduces performance, showing that principled evidence selection is important for decision-relevant reasoning.

\section{Conclusion}

In this paper, we presented BrainCoT, a multi-task brain signal foundation model for plug-and-play zero-shot inference with evidence-grounded reasoning. By unifying EEG/iEEG tasks into instruction-driven QA and introducing neurometric-anchored, decision-consistent reasoning, BrainCoT enables reliable inference without target-data-specific adaptation. Experiments on 12 datasets show strong zero-shot performance and improved direct decoding with verifiable intermediate evidence.

\begin{ack}

This work has been supported by the New Cornerstone Science Foundation. This work has been supported by Guangdong Provincial Key Laboratory of Ultra High Definition Immersive Media Technology (Grant No.
2024B1212010006).

The authors declare no competing interests.

\end{ack}

{
\small
\bibliographystyle{unsrt}
\bibliography{reference}
}

\appendix
\appendixtoc

\let\section\appsection
\let\subsection\appsubsection
\let\subsubsection\appsubsubsection

\section{Related Works}
\label{app:related_work}

\subsection{Brain Signal Foundation Models}

Many brain signal foundation models learn general-purpose representations via large-scale pretraining and evaluate them through downstream adaptation.
Representative examples include BrainBERT \cite{brainbert} for iEEG masked modeling and BIOT \cite{biot} for heterogeneous biosignals via unified tokenization to handle channel mismatch and missing values. Recent EEG-specific foundation models further introduce structure-aware designs, such as patch-based modeling and neural tokenization in LaBraM \cite{labram}, spatiotemporal modeling with format-adaptive encoding in CBraMod \cite{cbramod}, and cross-scale structure learning in CSBrain \cite{csbrain}. In parallel, BrainWave \cite{brainwave} targets clinical robustness by pretraining across scalp and intracranial EEG with broad evaluations. CodeBrain \cite{codebrain} combines decoupled temporal--frequency tokenization with a multi-scale architecture to capture local and long-range dependencies. SPOTR \cite{spotr} explores self-supervised physiological signal learning through reconstruction from a single-token global bottleneck. These approaches advance transferable representations, while representation transfer through downstream adaptation remains distinct from direct zero-shot inference on unseen datasets.

Multi-task EEG classification has also been explored through architectural designs. EEGMamba \cite{eegmamba} integrates spatiotemporal-adaptive processing, bidirectional state-space modeling, and task-aware mixture-of-experts routing.
To unify tasks, recent work introduces language or instruction tuning for multi-task EEG decoding. NeuroLM \cite{neurolm} and UniMind \cite{unimind} connect EEG with language under unified prompting, enabling a single model to support multiple tasks, although multi-task capability alone does not establish zero-shot transfer without target-specific adaptation.

Recent work explores signal--language alignment to connect brain signals with semantic descriptions and task instructions. ELM-MIL \cite{elmmil} aligns EEG with clinical reports using multiple-instance learning to handle weak correspondence, showing label-efficient and partial zero-shot capability. SSDDB \cite{ssddb} combines structural self-distillation with cross-modal contrastive alignment and evaluates its representations under full fine-tuning and few-shot settings. LEGEND \cite{legend} combines signal--language alignment with flow-matching latent denoising and evaluates zero-shot classification on unseen datasets. METIS \cite{metis} develops a unified language--signal alignment framework for multi-task zero-shot brain signal analysis. These approaches differ in their training objectives and adaptation requirements, but share an emphasis on semantic alignment for brain signal analysis. Semantic alignment alone, however, does not ensure that predictions are accompanied by externally checkable reasoning grounded in the input signals. BrainCoT addresses this additional requirement through neurometric-anchored reasoning and decision-consistent rationale training.

\subsection{Chain-of-Thought for Time-series}

Recent work on time-series reasoning often uses large language models (LLMs) to generate chain-of-thought (CoT) rationales to boost downstream performance. OpenTSLM \cite{opentslm} follows an LLM-as-annotator paradigm: it visualizes time series as plots and uses GPT-4o to produce CoT rationales conditioned on the gold label. However, such rationales are closer to post-hoc explanations and are not explicitly grounded in signal evidence, making reliability hard to guarantee. Similar limitations appear in GPT-supervised CoT fine-tuning and prompt-based step-by-step elicitation \cite{ts_reasoning_1,ts_reasoning_2}, where the resulting reasoning chains are typically not verifiable. To mitigate CoT errors, a generator–reviewer LLM setup has been introduced to revise reasoning chains \cite{ts_reasoning_3}, yet revisions can still reflect the reviewer’s linguistic preferences rather than true evidence alignment. Another direction, COUNTS \cite{ts_reasoning_4}, applies reinforcement learning with verifiable task metrics to induce explicit CoT; nevertheless, when optimization emphasizes final-answer rewards, models may still exhibit ``right answer, wrong reasoning''.

\newpage

\section{Brain Signal Instruction Dataset Construction Template}
\label{app_qa_template}

\newcolumntype{C}[1]{>{\Centering\arraybackslash}p{#1}}
\newcolumntype{L}[1]{>{\RaggedRight\arraybackslash}p{#1}}
\newcolumntype{Y}{>{\RaggedRight\arraybackslash}X}

\newcommand{\seprow}{\midrule[0.3pt]}

\newcommand{\qcell}[4]{%
\begin{minipage}[t]{\linewidth}
\RaggedRight
\textbf{T1} #1\par
\textbf{T2} #2\par
\textbf{T3} #3\par
\textbf{T4} #4
\end{minipage}
}

\newcommand{\acell}[1]{%
\begin{minipage}[t]{\linewidth}
\RaggedRight
#1
\end{minipage}
}

\begin{table}[ht]
\centering
\footnotesize
\setlength{\tabcolsep}{4pt}
\renewcommand{\arraystretch}{1.15}

\caption{\textbf{Pretraining datasets, task prompts, and canonical answer spaces.}
Each dataset uses four paraphrased question templates; multiple-choice questions are constructed by appending the option list (canonical answers) to the corresponding open-ended question.}
\label{tab:pretrain_qa_templates_all}

\begin{tabularx}{\textwidth}{@{}C{1.7cm}C{1.4cm}Y L{3.6cm}@{}}
\toprule
\textbf{Task} & \textbf{Dataset} & \multicolumn{1}{c}{\textbf{Question templates}} & \textbf{Canonical answers} \\
\midrule

Epileptic state recognition & HUP &
\qcell
{Which epilepsy state does this signal belong to?}
{Identify the epilepsy state of this recording segment.}
{Classify the epilepsy state for the given neural signal.}
{What epilepsy state best describes this signal?}
&
\acell{Interictal; Ictal; Preictal; Postictal} \\
\seprow

Epileptic state recognition & UPenn &
\qcell
{Which epilepsy state does this signal belong to?}
{Identify the epilepsy state of this recording segment.}
{Classify the epilepsy state for the given neural signal.}
{What epilepsy state best describes this signal?}
&
\acell{Interictal; Ictal} \\
\seprow

Epileptic state recognition & SWEC\_\allowbreak ETHZ &
\qcell
{Which epilepsy state does this signal belong to?}
{Identify the epilepsy state of this recording segment.}
{Classify the epilepsy state for the given neural signal.}
{What epilepsy state best describes this signal?}
&
\acell{Preictal; Ictal; Postictal} \\
\seprow

Interictal epileptiform discharge detection & FNUSA &
\qcell
{Does this signal contain interictal epileptiform activity?}
{Is interictal epileptiform activity present in this recording segment?}
{Determine whether this signal shows interictal epileptiform activity.}
{Does this signal show evidence of interictal epileptiform activity?}
&
\acell{Non-epileptiform activity; Interictal epileptiform activity} \\
\seprow

Epileptic state recognition & SeiZeIT2 &
\qcell
{Which epilepsy state does this signal belong to?}
{Identify the epilepsy state of this recording segment.}
{Classify the epilepsy state for the given neural signal.}
{What epilepsy state best describes this signal?}
&
\acell{Interictal; Ictal; Preictal; Postictal} \\
\seprow

Epileptic state recognition & CHB-MIT &
\qcell
{Which epilepsy state does this signal belong to?}
{Identify the epilepsy state of this recording segment.}
{Classify the epilepsy state for the given neural signal.}
{What epilepsy state best describes this signal?}
&
\acell{Interictal; Ictal; Preictal; Postictal} \\
\seprow

Sleep stage classification & SHHS &
\qcell
{Which sleep stage does this signal belong to?}
{Which sleep stage best describes this signal?}
{Identify the sleep stage of this signal.}
{Classify the sleep stage for this recording segment.}
&
\acell{Wake; Non-REM Stage 1; Non-REM Stage 2; Non-REM Stage 3; Rapid Eye Movement} \\
\seprow

Sleep stage classification & HMC &
\qcell
{Which sleep stage does this signal belong to?}
{Which sleep stage best describes this signal?}
{Identify the sleep stage of this signal.}
{Classify the sleep stage for this recording segment.}
&
\acell{Wake; Non-REM Stage 1; Non-REM Stage 2; Non-REM Stage 3; Rapid Eye Movement} \\

\bottomrule
\end{tabularx}

\vspace{-0.2cm}
\end{table}

\clearpage

\begin{center}
\footnotesize
\setlength{\tabcolsep}{4pt}
\renewcommand{\arraystretch}{1.10}

\captionsetup{type=table}
\caption{\textbf{Pretraining datasets, task prompts, and canonical answer spaces (continued).}}
\label{tab:pretrain_qa_templates_all_cont}

\vspace{0.3em}

\begin{tabularx}{\textwidth}{@{}C{1.3cm}C{1cm}Y L{4cm}@{}}
\toprule
\textbf{Task} & \textbf{Dataset} & \multicolumn{1}{c}{\textbf{Question templates}} & \textbf{Canonical answers} \\
\midrule

Anomaly detection & TUAB &
\qcell
{Is the signal normal or abnormal?}
{Determine whether this EEG segment is normal or abnormal.}
{Classify this EEG recording by its clinical status.}
{Does this signal indicate an abnormal EEG pattern?}
&
\acell{Normal; Abnormal} \\
\seprow

Clinical diagnosis classification & TUEP &
\qcell
{What clinical diagnosis is associated with this signal?}
{Which clinical condition label applies to this signal?}
{Assign the clinical condition for this recording segment.}
{Determine the clinical condition indicated by this signal.}
&
\acell{No epilepsy; Epilepsy} \\
\seprow

EEG event type classification & TUEV &
\qcell
{What EEG event type is present in this segment?}
{Identify the EEG event type for this signal.}
{Classify this segment by EEG event type.}
{Which EEG pattern/event label applies to this segment?}
&
\acell{Spike and slow wave; Generalized periodic epileptiform discharge; Periodic lateralized epileptiform discharge; Eye movement; Artifact; Background} \\
\seprow

Seizure type classification & TUSZ &
\qcell
{Which seizure type does this signal belong to?}
{Identify the seizure type of this recording segment.}
{Classify the seizure type for the given neural signal.}
{What seizure type best describes this signal?}
&
\acell{Eight seizure subtypes. See Table~\ref{tab:tusz_canonical_answers}.} \\
\seprow

Clinical diagnosis classification & TDBrain &
\qcell
{What clinical diagnosis is associated with this signal?}
{Which clinical condition label applies to this signal?}
{Assign the clinical condition for this recording segment.}
{Determine the clinical condition indicated by this signal.}
&
\acell{Dataset-specific clinical labels. See Table~\ref{tab:tdbrain_canonical_answers}.} \\
\seprow

Clinical diagnosis classification & MDD &
\qcell
{What clinical diagnosis is associated with this signal?}
{Which clinical condition label applies to this signal?}
{Assign the clinical condition for this recording segment.}
{Determine the clinical condition indicated by this signal.}
&
\acell{Normal; Major depressive disorder} \\
\seprow

Clinical diagnosis classification & ADFTD &
\qcell
{What clinical diagnosis is associated with this signal?}
{Which clinical condition label applies to this signal?}
{Assign the clinical condition for this recording segment.}
{Determine the clinical condition indicated by this signal.}
&
\acell{Normal; Alzheimer's disease; Frontotemporal dementia} \\
\seprow

Clinical diagnosis classification & BrainLat &
\qcell
{What clinical diagnosis is associated with this signal?}
{Which clinical condition label applies to this signal?}
{Assign the clinical condition for this recording segment.}
{Determine the clinical condition indicated by this signal.}
&
\acell{Normal; Alzheimer's disease; Frontotemporal dementia; Parkinson's disease; Multiple sclerosis} \\
\seprow

Clinical diagnosis classification & AD-Auditory &
\qcell
{What clinical diagnosis is associated with this signal?}
{Which clinical condition label applies to this signal?}
{Assign the clinical condition for this recording segment.}
{Determine the clinical condition indicated by this signal.}
&
\acell{Normal; Alzheimer's disease; Mild cognitive impairment} \\
\seprow

Clinical diagnosis classification & REEG-PD &
\qcell
{What clinical diagnosis is associated with this signal?}
{Which clinical condition label applies to this signal?}
{Assign the clinical condition for this recording segment.}
{Determine the clinical condition indicated by this signal?}
&
\acell{Normal; Parkinson's disease} \\

\bottomrule
\end{tabularx}

\end{center}

\newpage
\begin{center}
\small
\setlength{\tabcolsep}{6pt}
\renewcommand{\arraystretch}{1.4}

\begin{longtable}{L{0.08\textwidth} L{0.85\textwidth}}
\caption{TDBrain label inventory and canonical answer strings.}
\label{tab:tdbrain_canonical_answers}\\

\toprule
\textbf{Class} & \textbf{Canonical answer string} \\
\midrule
\endfirsthead

\multicolumn{2}{l}{\textit{Table~\thetable\ (continued).}} \\
\toprule
\textbf{Class} & \textbf{Canonical answer string} \\
\midrule
\endhead

\midrule
\multicolumn{2}{r}{\textit{(continued on next page)}} \\
\endfoot

\bottomrule
\endlastfoot

0  & "Burnout" \\
1  & "Subjective memory complaints" \\
2  & "Normal" \\
3  & "Dyslexia" \\
4  & "Chronic pain" \\
5  & "Major depressive disorder" \\
6  & "Attention deficit hyperactivity disorder" \\
7  & "Attention deficit hyperactivity disorder, asperger syndrome" \\
8  & "Pervasive developmental disorder not otherwise specified, dyslexia" \\
9  & "Pervasive developmental disorder not otherwise specified" \\
10 & "Whiplash" \\
11 & "Anxiety" \\
12 & "Attention deficit hyperactivity disorder, dyslexia" \\
13 & "Autism spectrum disorder" \\
14 & "Tinnitus" \\
15 & "Obsessive-compulsive disorder" \\
16 & "Panic disorder" \\
17 & "Major depressive disorder, anxiety" \\
18 & "Migraine" \\
19 & "Pervasive developmental disorder not otherwise specified, anxiety" \\
20 & "Parkinson's disease" \\
21 & "Bipolar disorder" \\
22 & "Major depressive disorder, bipolar disorder" \\
23 & "Dyspraxia" \\
24 & "Tinnitus, major depressive disorder" \\
25 & "Attention deficit hyperactivity disorder, autism spectrum disorder, anxiety" \\
26 & "Major depressive disorder, attention deficit hyperactivity disorder" \\
27 & "Attention deficit hyperactivity disorder, pervasive developmental disorder not otherwise specified" \\
28 & "Asperger syndrome" \\
29 & "Attention deficit hyperactivity disorder, epilepsy" \\
30 & "Major depressive disorder, pain" \\
31 & "Pervasive developmental disorder not otherwise specified, gilles de la tourette syndrome" \\
32 & "Pervasive developmental disorder not otherwise specified, attention deficit hyperactivity disorder" \\
33 & "Pervasive developmental disorder not otherwise specified, autism spectrum disorder" \\
34 & "Traumatic brain injury" \\
35 & "Attention deficit hyperactivity disorder, anxiety" \\
36 & "Attention deficit hyperactivity disorder, dyslexia, dyscalculia" \\
37 & "Attention deficit hyperactivity disorder, major depressive disorder" \\
38 & "Major depressive disorder, panic disorder" \\
39 & "Depersonalization disorder" \\
40 & "Major depressive disorder, trauma" \\
41 & "Post-traumatic stress disorder, attention deficit hyperactivity disorder" \\
42 & "Obsessive-compulsive disorder, dissociative psychogenic seizures" \\
43 & "Major depressive disorder, obsessive-compulsive disorder" \\
44 & "Major depressive disorder, tumor" \\
45 & "Attention deficit hyperactivity disorder, gilles de la tourette syndrome" \\
46 & "Obsessive-compulsive disorder, major depressive disorder" \\
47 & "Conversion disorder" \\
48 & "Autism spectrum disorder, asperger syndrome" \\
49 & "Major depressive disorder, attention deficit hyperactivity disorder, lyme disease" \\
50 & "Attention deficit hyperactivity disorder, obsessive-compulsive disorder" \\
51 & "Multiple system atrophy - cerebellar type" \\
52 & "Obsessive-compulsive disorder, autism spectrum disorder" \\
53 & "Stroke, pain" \\
54 & "Stroke" \\
55 & "Major depressive disorder, obsessive-compulsive disorder, attention deficit hyperactivity disorder" \\
56 & "Epilepsy, obsessive-compulsive disorder" \\
57 & "Insomnia" \\
58 & "Major depressive disorder, attention deficit hyperactivity disorder, anorexia" \\
59 & "Major depressive disorder, anxiety, tinnitus" \\

\end{longtable}
\end{center}

\begin{center}
\small
\setlength{\tabcolsep}{6pt}
\renewcommand{\arraystretch}{1.4}

\begin{longtable}{L{0.08\textwidth} L{0.85\textwidth}}
\caption{TUSZ label inventory and canonical answer strings.}
\label{tab:tusz_canonical_answers}\\

\toprule
\textbf{Class} & \textbf{Canonical answer string} \\
\midrule
\endfirsthead

\multicolumn{2}{l}{\textit{Table~\thetable\ (continued).}} \\
\toprule
\textbf{Class} & \textbf{Canonical answer string} \\
\midrule
\endhead

\midrule
\multicolumn{2}{r}{\textit{(continued on next page)}} \\
\endfoot

\bottomrule
\endlastfoot

0  & "Focal nonspecific seizure" \\
1  & "Generalized nonspecific seizure" \\
2  & "Simple partial seizure" \\
3  & "Complex partial seizure" \\
4  & "Absence seizure" \\
5  & "Tonic seizure" \\
6  & "Tonic-clonic seizure" \\
7  & "Myoclonic seizure" \\

\end{longtable}
\end{center}

\newpage
\section{Neurometric Evidence Extraction}
\label{app_neurometric}

We extract a deterministic set of \emph{neurometric features} from each brain signal segment
$X\in\mathbb{R}^{C\times L}$ to provide reproducible evidence for neurometric anchoring.
We first apply per-channel $z$-score normalization:
\begin{equation}
\tilde{x}_c[t]=\frac{x_c[t]-\mu_c}{\sigma_c+\varepsilon},
\qquad
\mu_c=\frac{1}{L}\sum_{t=1}^{L}x_c[t],
\qquad
\sigma_c=\sqrt{\frac{1}{L}\sum_{t=1}^{L}\bigl(x_c[t]-\mu_c\bigr)^2},
\end{equation}
where $\varepsilon$ is a small constant for numerical stability.
For each channel, we estimate the one-sided power spectral density (PSD) $P_c(f)$ using Welch's method
(with a Hann window, 50\% overlap, and constant detrending).
Based on $P_c(f)$, we compute 12 neurometric features per channel, covering three categories
(\emph{Spectral Structure Analysis}, \emph{Band Power Analysis}, and \emph{Time Domain Analysis}).

\textbf{Spectral Structure Analysis.}
We compute \emph{Spectral Entropy} over 0.5--45 Hz:
\begin{equation}
\mathrm{SE}_c
=-\frac{1}{\log N}\sum_{i=1}^{N}p_i\log p_i,
\qquad
p_i=\frac{P_c(f_i)}{\sum_{j=1}^{N}P_c(f_j)+\varepsilon},
\end{equation}
where $\{f_i\}_{i=1}^{N}$ are the PSD bins within 0.5--45 Hz.
We further estimate the \emph{Aperiodic Spectral Slope} via log--log linear regression within 2--40 Hz:
\begin{equation}
\mathrm{Slope}_c = a,
\quad
\text{where } (a,b)=\arg\min_{a,b}\sum_{f_i\in[2,40]}
\Bigl(\log_{10}(P_c(f_i)+\varepsilon)-a\log_{10}(f_i+\varepsilon)-b\Bigr)^2.
\end{equation}

\textbf{Band Power Analysis.}
We compute the \emph{Relative $\delta$ Power}, \emph{Relative $\theta$ Power}, \emph{Relative $\alpha$ Power},
\emph{Relative $\beta$ Power}, and \emph{Relative $\gamma$ Power} as:
\begin{equation}
\mathrm{RelBP}_c([a,b])=
\frac{\int_{a}^{b} P_c(f)\,df}{\int_{0.5}^{45} P_c(f)\,df+\varepsilon}.
\end{equation}

\textbf{Time Domain Analysis.}
We compute \emph{Line Length} as the mean absolute first difference,
\begin{equation}
\mathrm{LL}_c=\frac{1}{L-1}\sum_{t=1}^{L-1}\bigl|\tilde{x}_c[t+1]-\tilde{x}_c[t]\bigr|,
\end{equation}
as well as \emph{Hjorth Mobility} and \emph{Hjorth Complexity}:
\begin{equation}
\mathrm{Mob}_c=\sqrt{\frac{\mathrm{Var}(\Delta\tilde{x}_c)}{\mathrm{Var}(\tilde{x}_c)+\varepsilon}},
\qquad
\mathrm{Comp}_c=\frac{\sqrt{\frac{\mathrm{Var}(\Delta^2\tilde{x}_c)}{\mathrm{Var}(\Delta\tilde{x}_c)+\varepsilon}}}{\mathrm{Mob}_c+\varepsilon},
\end{equation}
where $\Delta\tilde{x}_c[t]=\tilde{x}_c[t+1]-\tilde{x}_c[t]$ and $\Delta^2\tilde{x}_c[t]=\Delta\tilde{x}_c[t+1]-\Delta\tilde{x}_c[t]$.
We further compute the \emph{Zero-Crossing Rate},
\begin{equation}
\mathrm{ZCR}_c=\frac{1}{L-1}\sum_{t=1}^{L-1}\mathbb{I}\bigl(\tilde{x}_c[t]\cdot \tilde{x}_c[t+1]<0\bigr),
\end{equation}
and \emph{Pearson Kurtosis},
\begin{equation}
\mathrm{Kurt}_c=\frac{\mathbb{E}\bigl[(\tilde{x}_c-\mathbb{E}[\tilde{x}_c])^4\bigr]}
{\mathbb{E}\bigl[(\tilde{x}_c-\mathbb{E}[\tilde{x}_c])^2\bigr]^2+\varepsilon}.
\end{equation}

Finally, to obtain a fixed-length descriptor for segments with varying channel counts, we compute the above features
independently for each channel and aggregate them across channels using four statistics: mean, standard deviation,
median, and interquartile range (IQR). Concatenating these aggregated statistics yields the final neurometric feature
vector $\phi(X)\in\mathbb{R}^{48}$ for each segment.

\newpage
\section{Reasoning-trajectory Templates}
\label{app_reason_trace_template}

To provide verifiable and reproducible supervision for \textit{Reasoning Mode}, we construct each reasoning trajectory using a compact template system grounded in deterministic neurometric computations. Each trajectory is built from three ingredients: (i) a small set of neurometric evidence items extracted from the input segment, (ii) a discretized \emph{level} descriptor for each evidence item based on within-dataset quantiles, and (iii) a natural-language sentence template that verbalizes the selected evidence. This design avoids free-form narrative generation and ensures that every statement in the trajectory corresponds to a reproducible computation on the input signal.

\textbf{Neurometric evidence pool.}
Given a signal segment $X$, we compute a fixed pool of neurometric features. Briefly, we extract 12 base measures covering \emph{Spectral Structure Analysis} (Spectral Entropy, Aperiodic Spectral Slope), \emph{Band Power Analysis} (Relative $\delta/\theta/\alpha/\beta/\gamma$ Power), and \emph{Time Domain Analysis} (Line Length, Hjorth Mobility, Hjorth Complexity, Zero-Crossing Rate, Pearson Kurtosis). For multi-channel recordings, each base measure is computed per channel and then aggregated across channels using four statistics: mean, standard deviation, median, and interquartile range (IQR), resulting in $12\times4=48$ candidate evidence features per segment. In the reasoning trajectory text, neurometric evidence is expressed using descriptive feature names (e.g., "Mean relative alpha power is relatively high").

\textbf{Quantile-based level mapping.}
To avoid brittle dependence on absolute numeric values, we convert each selected feature value into a discrete level based on within-dataset empirical quantiles, so evidence is expressed as relative strength (e.g., low/high). For feature dimension $j$, we compute four quantile thresholds on the training split of the same dataset:
\[
\begin{aligned}
q_j^{(1)} &= \operatorname{Quantile}(F_j, 0.2),
&\qquad q_j^{(2)} &= \operatorname{Quantile}(F_j, 0.4), \\
q_j^{(3)} &= \operatorname{Quantile}(F_j, 0.6),
&\qquad q_j^{(4)} &= \operatorname{Quantile}(F_j, 0.8).
\end{aligned}
\]
Given the feature value $F_j(X)$ for a segment, we assign it to one of five bins:
\begin{equation}
\ell_j(X)=
\begin{cases}
0, & F_j(X) < q_j^{(1)}\\
1, & q_j^{(1)} \le F_j(X) < q_j^{(2)}\\
2, & q_j^{(2)} \le F_j(X) < q_j^{(3)}\\
3, & q_j^{(3)} \le F_j(X) < q_j^{(4)}\\
4, & F_j(X) \ge q_j^{(4)}.
\end{cases}
\end{equation}
We map $\ell \in \{0,1,2,3,4\}$ to five fixed descriptors:
low, relatively low, moderate, relatively high, and high.
Each evidence item is then verbalized as "\emph{(feature name) is (level descriptor)}", e.g., "Mean spectral entropy is relatively low".

\textbf{Template rendering.}
A trajectory contains $K$ evidence items (we use $K=4$ in our implementation). To introduce mild linguistic diversity without changing the underlying evidence, we render the $K$ evidence phrases into a single sentence using one of the following templates sampled uniformly at random:
\begin{itemize}
  \item The signal shows that \texttt{\{F1\}}; additionally, \texttt{\{F2\}}, \texttt{\{F3\}}, and \texttt{\{F4\}}.
  \item In this segment, \texttt{\{F1\}}; furthermore, \texttt{\{F2\}}, \texttt{\{F3\}}, and \texttt{\{F4\}}.
  \item The signal indicates that \texttt{\{F1\}}, and that \texttt{\{F2\}}, \texttt{\{F3\}}, and \texttt{\{F4\}}.
  \item The signal presents the following patterns: \texttt{\{F1\}}, \texttt{\{F2\}}, \texttt{\{F3\}}, and \texttt{\{F4\}}.
\end{itemize}
Here, $\{F1\}\ldots\{F4\}$ denote the four evidence phrases produced by the quantile-based mapping above (feature selection follows the MI-guided strategy described in the main text). Because each phrase is deterministically derived from neurometric computations on $X$ given dataset-specific quantile thresholds, the resulting reasoning trajectory is reproducible and verifiable, and can be used as reliable supervision for reasoning training.

The use of structured templates for reasoning trajectories is a deliberate design choice to prioritize logical faithfulness and clinical verifiability over surface-level linguistic diversity. In high-stakes medical domains, unconstrained free-form Chain-of-Thought (CoT)~\cite{cot} is often susceptible to 'hallucinated evidence' that lacks physiological grounding. By constraining the intermediate reasoning steps to deterministic neurometric features, we ensure that the model’s 'thought process' is externally checkable and reproducible. This approach is conceptually consistent with recent 'step-level verification' paradigms~\cite{verifycot} and rule-governed reasoning frameworks~\cite{deepseekr1}, which demonstrate that structured constraints can effectively incentivize more rigorous and non-hallucinatory decision-making in complex tasks.

\newpage
\section{Model Architecture}
\label{app_model_arch}
\begin{figure*}[htb]
\begin{center}
\centerline{\includegraphics[width=\textwidth]{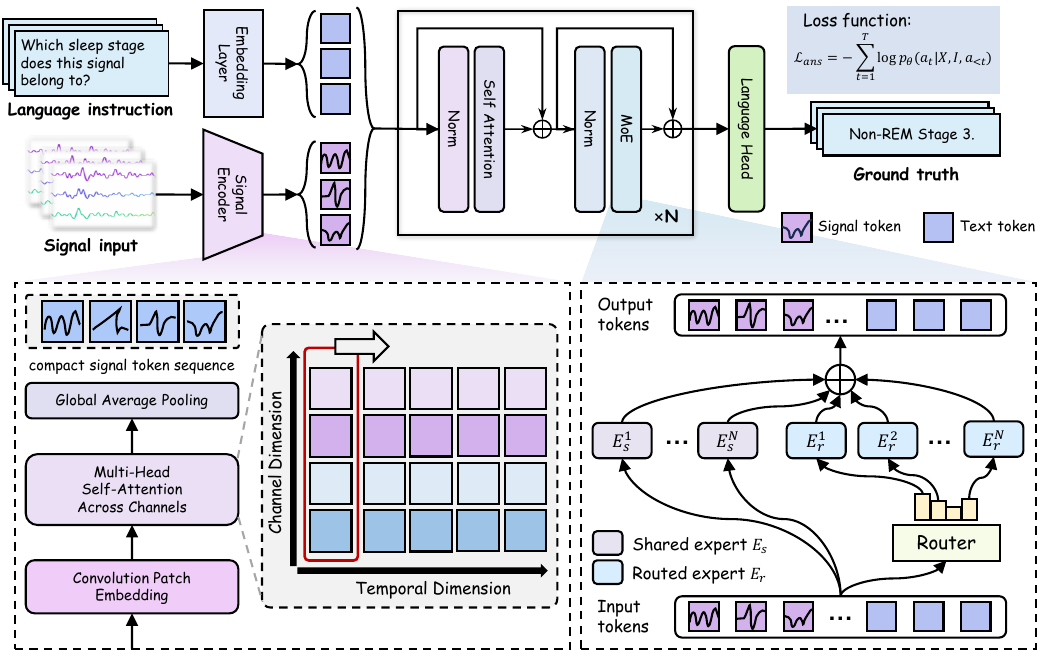}}
\caption{\textbf{The model architecture of BrainCoT.}}
\label{fig4_model_architecture}
\end{center}
\vspace{-0.5cm}
\end{figure*}

BrainCoT follows a unified decoder-only Transformer architecture for instruction-driven brain signal modeling, consisting of (i) a universal signal encoder that converts heterogeneous EEG/iEEG signals into a unified token space, and (ii) a Mixture-of-Experts (MoE) Transformer backbone that performs adaptive computation conditioned on token content and task instruction.

\paragraph{Universal signal encoder}
To enable a single backbone to process brain signals from different datasets and acquisition systems, we first map heterogeneous inputs into a shared feature space. Given a raw signal
$x \in \mathbb{R}^{B \times C \times L}$ with batch size $B$, channel number $C$, and sequence length $L$, we apply channel-wise $z$-score normalization to reduce scale variation across subjects/devices and stabilize optimization. We then use channel-independent convolutional patch-embedding layers $\mathcal{A}(\cdot)$ to tokenize the signal along the temporal axis:
\begin{equation}
U = \mathcal{A}(x) \in \mathbb{R}^{B \times C \times N \times D},
\end{equation}
where $D$ is the hidden dimension and $N$ is the number of temporal tokens per channel.

To obtain a compact representation and reduce subsequent computation, we compress tokens along the channel dimension via an attention pooling operator. Specifically, for each temporal index $i \in \{1,\dots,N\}$, we aggregate channel tokens $U_{:,:,i,:}$ using multi-head attention followed by a pooling function:
\begin{equation}
z_i = \mathrm{Pool}\!\left(\mathrm{MHA}\!\left(U_{:,:,i,:}\right)\right)\in \mathbb{R}^{B \times D}, \quad i=1,2,\dots,N,
\end{equation}
yielding the final signal-token sequence $Z_s=\{z_1,z_2,\dots,z_N\}\in \mathbb{R}^{B \times N \times D}$.

We then form the backbone input by concatenating the signal-token sequence with the instruction tokens.
Let $Z_s=\{z^{(s)}_1,\dots,z^{(s)}_N\}\in\mathbb{R}^{B\times N\times D}$ denote the signal-token sequence from the encoder, and
let $Z_t=\{z^{(t)}_1,\dots,z^{(t)}_M\}\in\mathbb{R}^{B\times M\times D}$ denote the token embeddings of the textual instruction produced by the language tokenizer and embedding layer.
The final input to the Transformer backbone is the concatenation
\begin{equation}
Z = [Z_s; Z_t]\in\mathbb{R}^{B\times (N+M)\times D},
\end{equation}
which is fed into the decoder-only backbone to autoregressively generate reasoning and answer tokens.

\paragraph{Mixture-of-Experts backbone}
To enhance task-specific and modality-specific processing, we replace the standard feed-forward network (FFN) in Transformer blocks with an MoE layer. For an input token representation $x \in \mathbb{R}^{D}$, a router $g(\cdot)$ produces gating logits
\begin{equation}
g(x) = W_g x,\quad W_g \in \mathbb{R}^{N_e \times D},
\end{equation}
where $N_e$ denotes the number of routed experts. The routing distribution is
\begin{equation}
p(x) = \mathrm{softmax}(g(x)).
\end{equation}
We adopt top-$k$ routing to activate the $k$ experts with the largest routing probabilities, and additionally include a shared expert to capture common structure across tasks/modalities and improve training stability. Let $E_j(\cdot)$ be the $j$-th routed expert, $E_s(\cdot)$ the shared expert, and $m_j(x)\in\{0,1\}$ the top-$k$ selection mask. The MoE output is
\begin{equation}
y(x) = \sum_{j=1}^{N_e} m_j(x)\, p_j(x)\, E_j(x) + E_s(x).
\end{equation}

To encourage balanced expert utilization, we add an auxiliary load-balancing objective:
\begin{equation}
\mathcal{L}_{\mathrm{aux}}=\sum_{e=1}^{N_e} p_e\, q_e,
\end{equation}
where $p_e$ is the fraction of tokens routed to expert $e$, and $q_e$ is the average routing probability assigned to expert $e$.

\newpage
\section{Implementation Details}

\textbf{Mutual information estimation.}
\label{MI_stimation}
To quantify the label relevance of each neurometric feature, we estimate the mutual information (MI) between feature values and class labels on each dataset. Given a feature matrix $X\in\mathbb{R}^{N\times D}$ and corresponding labels $y$, we compute a scalar MI score for each feature dimension $j$ as
\[
\mathrm{MI}_j = I(X_{\cdot j}; y),
\]
where $X_{\cdot j}$ denotes the $j$-th feature across samples. We employ a nonparametric $k$-nearest-neighbor mutual information estimator for continuous variables, as implemented in the \texttt{mutual\_info\_classif} function from the \texttt{scikit-learn} library, with $k{=}10$ neighbors and a fixed random seed for reproducibility. Before MI estimation, all feature values are converted to floating-point format and any NaN or infinite entries are replaced by zeros to ensure numerical stability. The resulting MI scores serve as relative relevance measures for neurometric feature selection in the reasoning-trajectory construction.

\textbf{Model configuration.}
Table~\ref{tab:impl_details} summarizes the architectural hyperparameters of BrainCoT.

\begin{table}[htb]
\centering
\caption{\textbf{Implementation details of BrainCoT.}}
\label{tab:impl_details}
\small
\setlength{\tabcolsep}{10pt}
\renewcommand{\arraystretch}{1.1}
\begin{tabular}{@{}l l@{}}
\toprule
\textbf{Hyper-parameter} & \textbf{Value} \\
\midrule
Transformer layers & 12 \\
Hidden dimension & 768 \\
FFN dimension & 3072 \\
Attention heads & 12 \\
Key--value heads & 2 \\
RoPE base & $10^{6}$ \\
Maximum sequence length & 2048 \\
Vocabulary size & 151{,}936 \\
\midrule
MoE layers & 10 \\
Dense FFN layers & 2 \\
Routed experts & 8 \\
Top-$k$ routing & 2 \\
Auxiliary loss weight & $10^{-4}$ \\
\midrule
Signal tokenizer & 2-layer convolutional \\
Patch stride & 100 \\
Channel attention & Non-causal \\
\bottomrule
\end{tabular}
\end{table}
For training stability, we use dense FFNs in the first two Transformer layers and apply MoE blocks in the remaining layers, following the practice of DeepSeek-V3~\cite{deepseekv3}.
The input embedding matrix and the language modeling head share the same parameters. For text tokenization, we use the Qwen3~\cite{Qwen3} tokenizer, whose large vocabulary helps reduce instruction length.
Overall, BrainCoT contains approximately 329.2M total parameters, while only about 223.0M parameters are activated per forward pass due to sparse expert routing.

\newpage
\section{Datasets}

\subsection{Pretraining Datasets}
\label{app_pretraining_datasets}

\begin{table}[htb]
\centering
\caption{\textbf{Pretraining datasets and corpus statistics.}}
\resizebox{0.75\textwidth}{!}{%
\begin{tabular}{lcccc}
\toprule
\textbf{Dataset} & \textbf{Data type} & \textbf{Subject} & \textbf{Samples} & \textbf{Duration (hours)} \\
\midrule
HUP \cite{hup} & iEEG & 57 & 3,322,242 & 2768.54 \\
UPenn \cite{upenn} & iEEG & 12 & 1,691,209 & 469.78 \\
SWEC\_ETHZ \cite{swecethz} & iEEG & 16 & 1,004,999 & 837.50 \\
FNUSA \cite{mayo} & iEEG & 12 & 147,030 & 122.53 \\
SHHS \cite{shhs} & EEG & 6,441 & 5,789,088 & 48,242.40 \\
SeiZeIT2 \cite{seizeit2} & EEG & 125 & 4,519,290 & 12,553.58 \\
TUSZ \cite{tusz} & EEG & 675 & 687,375 & 1,909.38 \\
TUAB \cite{tuab} & EEG & 2,329 & 409,455 & 1,137.38 \\
CHB-MIT \cite{chbmit} & EEG & 23 & 384,834 & 1,068.98 \\
TUEP \cite{tuep} & EEG & 179 & 214,376 & 595.49 \\
HMC \cite{hmc} & EEG & 151 & 137,244 & 1,143.70 \\
TDBrain \cite{tdbrain} & EEG & 911 & 115,668 & 64.26 \\
TUEV \cite{tuev} & EEG & 370 & 111,547 & 154.93 \\
ADFTD \cite{adftd} & EEG & 88 & 34,876 & 19.38 \\
BrainLat \cite{brainlat} & EEG & 135 & 30,699 & 17.06 \\
AD-Auditory \cite{ad-auditory} & EEG & 35 & 17,757 & 9.87 \\
REEG-PD \cite{reeg-pd} & EEG & 149 & 11,878 & 6.60 \\
MDD \cite{mdd} & EEG & 61 & 7,600 & 10.56 \\
\bottomrule
\end{tabular}
}
\label{tab:pretraining_datasets}
\end{table}

We curate a large-scale brain signal pretraining corpus spanning \textbf{18 datasets}, including \textbf{4 iEEG} datasets and \textbf{14 EEG} datasets (Table~\ref{tab:pretraining_datasets}). In total, the corpus aggregates recordings from \textbf{11,769 subjects}, comprising \textbf{18,637,167} segmented samples and \textbf{71,131.92} hours of signal data. This collection covers diverse acquisition settings, channel configurations, and clinical/physiological scenarios, providing broad supervision for learning transferable brain signal representations.

\newpage
\subsection{Downstream Datasets}
\label{app_downstream_datasets}
We summarize the downstream benchmarks below and cast all tasks into a unified instruction-driven QA form using the templates in Table~\ref{tab:downstream_qa_templates_all}.
We report \textbf{multiple-choice QA} for quantitative evaluation, and additionally include \textbf{open-ended QA} for qualitative visualization. Multiple-choice questions are formed by appending the option list (canonical answers) to the corresponding open-ended template. 

\textbf{ISRUC} \cite{isruc}: This sleep-stage classification benchmark includes overnight scalp EEG from 100 subjects (200 Hz). We adopt a two-channel montage (C3--M2 and C4--M1) and segment the recordings into 30-second epochs annotated with the five standard sleep stages, resulting in 73,883 labeled segments.

\textbf{Dreams} \cite{dreams}: This dataset provides overnight EEG from 20 subjects sampled at 200 Hz for sleep-stage classification. We use three channels (FP1--A2, FP2--A1, and CZ--A1) and construct 30-second epochs with five-stage sleep labels, yielding 20,242 segments.

\textbf{SEE} \cite{see}: SEE is an epilepsy dataset consisting of single-channel EEG from 30 subjects recorded at 173.61 Hz with a predefined train/test split. We crop the signals into 4-second windows and assign binary labels (epilepsy vs. no epilepsy), producing 4,000 segments in total.

\textbf{Siena} \cite{siena}: Siena contains multichannel scalp EEG from 14 subjects sampled at 512 Hz, annotated into four peri-ictal categories (pre-ictal, ictal, post-ictal, and inter-ictal). We use 16 channels and 10-second windows, obtaining 50,749 segments to evaluate seizure-related dynamics.

\textbf{IEDS} \cite{ieds}: IEDS is an iEEG dataset targeting interictal epileptiform activity, with recordings from 25 subjects at 1000 Hz. After segmenting into 3-second windows, each sample is labeled as normal activity or IED, resulting in 59,615 segments.

\textbf{Mayo} \cite{mayo}: The Mayo iEEG dataset includes recordings from 15 subjects at 500 Hz with four original labels. We exclude artifact- and power-line-related segments and formulate a binary interictal epileptiform activity (IEA) detection task. Using 3-second windows, we obtain 71,957 labeled segments.

\textbf{ADFSU} \cite{adfsu}: This Alzheimer’s disease dataset consists of 19-channel EEG recorded at 128 Hz from 92 participants (80 patients and 12 controls). We form 2-second samples with binary labels (AD vs.\ control), resulting in 736 segments.

\textbf{APAVA} \cite{apava}: APAVA is another Alzheimer’s disease dataset with 16-channel EEG at 256 Hz from 23 participants (12 patients and 11 controls). We segment the recordings into 2-second windows, producing 2,652 labeled samples.

\textbf{SanDiego} \cite{sandiego}: For Parkinson’s disease detection, SanDiego provides 32-channel EEG recorded at 512 Hz from 31 participants (15 patients and 16 controls). We use 2-second windows to construct 4,521 samples for binary classification.

\textbf{ADHD} \cite{adhd-121}: This dataset contains 19-channel EEG recorded at 128 Hz from 121 children for ADHD detection. We extract 5-second windows and assign binary labels (ADHD vs.\ control), yielding 3,322 segments.

\textbf{Schizo} \cite{schizo-youth}: This schizophrenia detection dataset includes 16-channel EEG sampled at 128 Hz from 84 participants (45 patients and 39 controls). We segment the signals into 5-second windows, obtaining 504 labeled samples.

\textbf{NMT} \cite{nmtab}: NMT is a large-scale EEG benchmark for anomaly detection with 16 channels at 200 Hz. Using the official split and 10-second windows, we construct 175,091 segments labeled as normal or abnormal to cover broad anomaly patterns.

We clarify that BrainCoT defines zero-shot generalization as the model's ability to perform inference on entirely unseen dataset distributions without any target-specific parameter updates or prior access to the target domain's labels. This definition aligns with the established protocol in cross-modal foundation models such as CLIP~\cite{clip}, where 'zero-shot' refers to generalizing learned semantic concepts to new data distributions rather than a total absence of task-related categories during pretraining. Since brain signals are highly sensitive to acquisition hardware and subject variability, achieving high performance on unseen datasets constitutes a rigorous test of representation robustness and instruction-following capability, distinguishing our approach from models that rely on transductive or few-shot adaptation.

\clearpage

\begin{center}
\footnotesize
\setlength{\tabcolsep}{4pt}
\renewcommand{\arraystretch}{1.08}

\captionsetup{type=table}
\caption{\textbf{Downstream datasets, task prompts, and canonical answer spaces.}
Each dataset uses four paraphrased question templates; multiple-choice questions are constructed by appending the option list (canonical answers) to the corresponding open-ended question.}
\label{tab:downstream_qa_templates_all}

\vspace{0.3em}

\begin{tabularx}{\textwidth}{@{}C{1.7cm}C{1cm}Y L{3.6cm}@{}}
\toprule
\textbf{Task} & \textbf{Dataset} & \multicolumn{1}{c}{\textbf{Question templates}} & \textbf{Canonical answers} \\
\midrule

Sleep stage classification & ISRUC &
\qcell
{Which sleep stage does this signal belong to?}
{Which sleep stage best describes this signal?}
{Identify the sleep stage of this signal.}
{Classify the sleep stage for this recording segment.}
&
\acell{Wake; Non-REM Stage 1; Non-REM Stage 2; Non-REM Stage 3; Rapid Eye Movement} \\
\seprow

Sleep stage classification & Dreams &
\qcell
{Which sleep stage does this signal belong to?}
{Which sleep stage best describes this signal?}
{Identify the sleep stage of this signal.}
{Classify the sleep stage for this recording segment.}
&
\acell{Wake; Non-REM Stage 1; Non-REM Stage 2; Non-REM Stage 3; Rapid Eye Movement} \\
\seprow

Epileptic state recognition & Siena &
\qcell
{Which epilepsy state does this signal belong to?}
{Identify the epilepsy state of this recording segment.}
{Classify the epilepsy state for the given neural signal.}
{What epilepsy state best describes this signal?}
&
\acell{Interictal; Ictal; Preictal; Postictal} \\
\seprow

Interictal epileptiform activity detection & Mayo &
\qcell
{Does this signal contain interictal epileptiform activity?}
{Is interictal epileptiform activity present in this recording segment?}
{Determine whether this signal shows interictal epileptiform activity.}
{Does this signal show evidence of interictal epileptiform activity?}
&
\acell{Non-epileptiform activity; Interictal epileptiform activity} \\
\seprow

Interictal epileptiform activity detection & IEDS &
\qcell
{Does this signal contain interictal epileptiform activity?}
{Is interictal epileptiform activity present in this recording segment?}
{Determine whether this signal shows interictal epileptiform activity.}
{Does this signal show evidence of interictal epileptiform activity?}
&
\acell{Non-epileptiform activity; Interictal epileptiform activity} \\
\seprow

Clinical diagnosis classification & SEE &
\qcell
{What clinical diagnosis is associated with this signal?}
{Which clinical condition label applies to this signal?}
{Assign the clinical condition for this recording segment.}
{Determine the clinical condition indicated by this signal.}
&
\acell{No epilepsy; Epilepsy} \\
\seprow

Anomaly detection & NMT &
\qcell
{Is the signal normal or abnormal?}
{Determine whether this EEG segment is normal or abnormal.}
{Classify this EEG recording by its clinical status.}
{Does this signal indicate an abnormal EEG pattern?}
&
\acell{Normal; Abnormal} \\
\seprow

Clinical diagnosis classification & ADHD &
\qcell
{What clinical diagnosis is associated with this signal?}
{Which clinical condition label applies to this signal?}
{Assign the clinical condition for this recording segment.}
{Determine the clinical condition indicated by this signal.}
&
\acell{Normal; Attention deficit hyperactivity disorder} \\
\seprow

Clinical diagnosis classification & Schizo &
\qcell
{What clinical diagnosis is associated with this signal?}
{Which clinical condition label applies to this signal?}
{Assign the clinical condition for this recording segment.}
{Determine the clinical condition indicated by this signal.}
&
\acell{Normal; Schizophrenia} \\
\seprow

Clinical diagnosis classification & ADFSU &
\qcell
{What clinical diagnosis is associated with this signal?}
{Which clinical condition label applies to this signal?}
{Assign the clinical condition for this recording segment.}
{Determine the clinical condition indicated by this signal.}
&
\acell{Normal; Alzheimer's disease} \\
\seprow

Clinical diagnosis classification & APAVA &
\qcell
{What clinical diagnosis is associated with this signal?}
{Which clinical condition label applies to this signal?}
{Assign the clinical condition for this recording segment.}
{Determine the clinical condition indicated by this signal.}
&
\acell{Normal; Alzheimer's disease} \\
\seprow

Clinical diagnosis classification & San\allowbreak Diego &
\qcell
{What clinical diagnosis is associated with this signal?}
{Which clinical condition label applies to this signal?}
{Assign the clinical condition for this recording segment.}
{Determine the clinical condition indicated by this signal.}
&
\acell{Normal; Parkinson's disease} \\

\bottomrule
\end{tabularx}

\end{center}

\newpage
\section{Baselines}
\label{app_baseline}

\textbf{BIOT} \cite{biot} is a generic transformer for biosignals (EEG, ECG, and wearable sensors). It tokenizes each channel into fixed-length segments, flattens them into a unified biosignal sentence, and adds channel and relative position embeddings before a linear-complexity transformer. For self-supervision, it uses channel/token dropout and contrastive loss. The raw code is available at \url{https://github.com/ycq091044/BIOT}.

\textbf{LaBraM} \cite{labram} is a foundation model for EEG signals. It segments EEG into channel patches, trains a vector-quantized neural tokenizer via neural spectrum prediction, then pretrains a Transformer by masked code prediction (BEiT/MAE-style) to learn transferable EEG representations across datasets and tasks. The raw code is available at \url{https://github.com/935963004/LaBraM}.

\textbf{CBraMod} \cite{cbramod} is a foundation model for EEG signals. It pretrains with patch-based masked EEG reconstruction using a time-frequency patch encoder, then applies a criss-cross transformer with separate spatial/temporal attentions and asymmetric conditional positional encoding to adapt across diverse channel layouts. The raw code is available at \url{https://github.com/wjq-learning/CBraMod}.

\textbf{CSBrain} \cite{csbrain} is a brain foundation model for EEG signals. It introduces cross-scale spatiotemporal tokenization to aggregate multi-scale features within local temporal windows and anatomical brain regions into scale-aware tokens, then applies structured sparse attention to model cross-window/region dependencies efficiently while suppressing spurious correlations. The raw code is available at \url{https://github.com/yuchen2199/CSBrain}.

\textbf{BrainBERT} \cite{brainbert} is a self-supervised representation learning framework for intracranial recordings. It adopts a masked autoencoding~\cite{mae} objective on time--frequency representations (e.g., spectrograms), reconstructing masked regions to learn transferable iEEG features from large-scale pretraining. The raw code is available at \url{https://github.com/czlwang/brainbert}.

\textbf{NeuroLM} \cite{neurolm} is a multi-task brain signal model built upon a pretrained language model (GPT-2 \cite{gpt2}) and a pretrained signal encoder. It jointly trains and evaluates on multiple downstream datasets under a unified prompting/interface, enabling a single model to perform multi-task brain signal analysis. The raw code is available at \url{https://github.com/935963004/NeuroLM}.

\textbf{ELM-MIL} \cite{elmmil} is a language--signal pretraining method for EEG abnormality detection. It introduces contrastive language--signal alignment and leverages a pretrained text encoder to embed label descriptions; after pretraining, the model supports zero-shot classification for EEG abnormality detection by matching signal embeddings with textual label embeddings. The raw code is available at \url{https://github.com/SamGijsen/ELM}.

\newpage
\section{Dataset-Level Zero-Shot Comparison with Linear-Probing Baselines}
\label{detail_zero-shot}
\begin{table}[!ht]
\caption{\textbf{Dataset-level comparison between BrainCoT zero-shot inference and 1\%/10\% linear-probing references.} \textbf{Bold} denotes the best performance. \textcolor{magenta}{Magenta} and \textcolor{teal}{teal} indicate whether BrainCoT zero-shot outperforms or underperforms the baseline, respectively. The color intensity is proportional to the absolute performance gap. The last column (“Zero-shot better than baselines”) counts baselines outperformed by zero-shot BrainCoT.}
\label{result_zero_shot}
\resizebox{\textwidth}{!}{%
\begin{tabular}{@{}cccccccccccc@{}}
\toprule\rowcolor{gray!15}
\multicolumn{3}{c}{\textbf{Model}} & \textbf{BrainBERT}   & \textbf{BIOT}        & \textbf{LaBraM }    & \textbf{CBraMod}     & \textbf{NeuroLM}     & \textbf{ELM-MIL}     & \textbf{CSBrain}    & \textbf{BrainCoT}    & {\cellcolor{magenta!0}Zero-shot better} \\ 
Dataset &   Metric      & Training data      & ICLR' 23           & NeurIPS' 23           & ICLR' 24           & ICLR' 25           & ICLR' 25           & ICML' 25           & NeurIPS' 25          & ours           &           than baselines                           \\ \midrule
                                     & & 0\% (zero-shot)                   &   -          &     -        &    -         &      -       &     -        &     -       &       -    & \textbf{82.77}$\pm{1.38}$  &                                      \\
                                     &\multirow{-1}{*}{ACC $\uparrow$} & 1\%                    & \cellcolor{magenta!17}65.45$\pm{2.06}$   & \cellcolor{magenta!40}43.27$\pm{6.71}$   & \cellcolor{magenta!45}37.95$\pm{4.79}$   & \cellcolor{magenta!48}35.09$\pm{6.07}$  & \cellcolor{magenta!44}38.60$\pm{4.02}$  & \cellcolor{magenta!46}37.01$\pm{6.81}$  & \cellcolor{magenta!57}25.68$\pm{1.30}$ & \textbf{84.07}$\pm{0.99}$    &     7 / 7\\
\multirow{-1}{*}[-0.4\baselineskip]{ISRUC}              & & 10\%                   & \cellcolor{magenta!9}73.80$\pm{1.83}$    & \cellcolor{magenta!23}59.60$\pm{14.74}$   & \cellcolor{magenta!29}53.36$\pm{8.33}$ & \cellcolor{magenta!48}35.23$\pm{11.93}$  & \cellcolor{magenta!25}57.50$\pm{8.28}$  & \cellcolor{magenta!24}58.61$\pm{9.22}$  & \cellcolor{magenta!52}30.47$\pm{0.79}$ & \textbf{85.26}$\pm{1.51}$   &     7 / 7\\ \cmidrule(lr){2-12}
& & 0\% (zero-shot)                   &   -          &     -        &    -         &      -       &     -        &     -       &       -    & \textbf{95.19}$\pm{0.62}$  &                                      \\
                                     &\multirow{-1}{*}{AUC $\uparrow$} & 1\%                    & \cellcolor{magenta!10}85.48$\pm{1.70}$   & \cellcolor{magenta!25}70.47$\pm{3.40}$   & \cellcolor{magenta!37}57.96$\pm{5.20}$   & \cellcolor{magenta!28}67.47$\pm{6.37}$  & \cellcolor{magenta!32}62.73$\pm{5.33}$  & \cellcolor{magenta!30}65.58$\pm{5.34}$  & \cellcolor{magenta!44}51.32$\pm{0.57}$ & \textbf{95.30}$\pm{0.40}$    &     7 / 7\\
& & 10\%                   & \cellcolor{magenta!5}90.40$\pm{1.10}$    & \cellcolor{magenta!10}85.65$\pm{5.30}$   & \cellcolor{magenta!26}69.19$\pm{10.87}$ & \cellcolor{magenta!22}73.08$\pm{9.54}$  & \cellcolor{magenta!17}77.88$\pm{7.94}$  & \cellcolor{magenta!14}80.94$\pm{8.32}$  & \cellcolor{magenta!40}55.39$\pm{2.89}$ & \textbf{96.20}$\pm{0.62}$   &     7 / 7\\ \midrule
                                     & & 0\% (zero-shot)                    &    -          &     -        &    -         &      -       &     -        &     -       &       -    & \textbf{77.66}$\pm{2.51}$  &                                      \\
                                     &\multirow{-1}{*}{ACC $\uparrow$}& 1\%                    & \cellcolor{magenta!30}47.29$\pm{3.91}$  & \cellcolor{magenta!35}42.43$\pm{5.72}$  & \cellcolor{magenta!25}53.02$\pm{5.05}$  & \cellcolor{magenta!32}45.48$\pm{3.89}$  & \cellcolor{magenta!40}37.98$\pm{0.78}$  & \cellcolor{magenta!37}40.46$\pm{1.63}$  & \cellcolor{magenta!41}36.83$\pm{9.41}$ & \textbf{77.09}$\pm{2.55}$  &     7 / 7\\
\multirow{-1}{*}[-0.4\baselineskip]{Dreams}             & & 10\%                   & \cellcolor{magenta!14}63.65$\pm{3.06}$  & \cellcolor{magenta!13}64.40$\pm{2.92}$  & \cellcolor{magenta!3}75.00$\pm{2.83}$  & \cellcolor{magenta!27}50.95$\pm{5.16}$  & \cellcolor{magenta!22}55.23$\pm{1.32}$  & \cellcolor{magenta!28}49.18$\pm{2.31}$  & \cellcolor{magenta!37}40.73$\pm{2.95}$ & \textbf{79.06}$\pm{1.85}$  &     7 / 7\\ \cmidrule(lr){2-12}

& & 0\% (zero-shot)                    &    -          &     -        &    -         &      -       &     -        &     -       &       -    & \textbf{92.49}$\pm{1.07}$  &                                      \\
                                     &\multirow{-1}{*}{AUC $\uparrow$}& 1\%                    & \cellcolor{magenta!17}75.53$\pm{3.32}$  & \cellcolor{magenta!21}71.81$\pm{8.63}$  & \cellcolor{magenta!17}75.10$\pm{3.16}$  & \cellcolor{magenta!27}65.53$\pm{5.04}$  & \cellcolor{magenta!30}62.25$\pm{1.37}$  & \cellcolor{magenta!32}60.93$\pm{3.84}$  & \cellcolor{magenta!42}50.63$\pm{0.68}$ & \textbf{92.17}$\pm{0.97}$  &     7 / 7\\
& & 10\%                   & \cellcolor{magenta!7}85.96$\pm{1.73}$  & \cellcolor{magenta!6}86.84$\pm{1.41}$  & \cellcolor{magenta!4}88.70$\pm{1.32}$  & \cellcolor{magenta!16}76.82$\pm{3.61}$  & \cellcolor{magenta!15}77.23$\pm{1.12}$  & \cellcolor{magenta!17}75.80$\pm{1.70}$  & \cellcolor{magenta!39}53.80$\pm{0.46}$ & \textbf{93.42}$\pm{0.77}$  &     7 / 7\\ \midrule
                                     & & 0\% (zero-shot)                    &        -          &     -        &    -         &      -       &     -        &     -       &       -    & \textbf{47.31}$\pm{16.20}$  &                                      \\
                                     &\multirow{-1}{*}{ACC $\uparrow$}& 1\%                    & \cellcolor{teal!4}51.40$\pm{14.57}$  & \cellcolor{teal!7}\textbf{53.93}$\pm{11.26}$  & \cellcolor{teal!3}49.88$\pm{11.62}$  & \cellcolor{teal!3}50.18$\pm{14.26}$  & \cellcolor{teal!3}50.80$\pm{12.69}$  & \cellcolor{teal!4}51.16$\pm{14.26}$  & \cellcolor{magenta!0}47.24$\pm{10.53}$ & 48.78$\pm{11.79}$   &     1 / 7\\
\multirow{-1}{*}[-0.4\baselineskip]{Siena}              & & 10\%                   & \cellcolor{teal!4}51.61$\pm{14.60}$  & \cellcolor{teal!4}51.64$\pm{14.55}$  & \cellcolor{teal!3}50.73$\pm{13.40}$   & \cellcolor{magenta!2}45.48$\pm{17.37}$   & \cellcolor{teal!1}47.81$\pm{11.67}$   & \cellcolor{teal!4}51.16$\pm{14.17}$  & \cellcolor{magenta!1}46.33$\pm{8.77}$ & \textbf{52.13}$\pm{9.87}$  &     2 / 7\\ \cmidrule(lr){2-12}
& & 0\% (zero-shot)                    &        -          &     -        &    -         &      -       &     -        &     -       &       -    & \textbf{64.85}$\pm{9.68}$  &                                      \\
                                     &\multirow{-1}{*}{AUC $\uparrow$}& 1\%                    & \cellcolor{magenta!7}58.08$\pm{3.99}$  & \cellcolor{magenta!7}57.47$\pm{3.82}$  & \cellcolor{magenta!8}56.62$\pm{1.76}$  & \cellcolor{magenta!24}40.82$\pm{2.75}$  & \cellcolor{magenta!13}52.12$\pm{1.02}$  & \cellcolor{magenta!5}59.44$\pm{2.84}$  & \cellcolor{magenta!12}52.47$\pm{2.69}$ & \textbf{62.77}$\pm{9.20}$   &     7 / 7\\
& & 10\%                   & \cellcolor{magenta!5}59.61$\pm{6.24}$  & \cellcolor{magenta!5}60.03$\pm{4.04}$  & \cellcolor{magenta!7}57.85$\pm{3.30}$   & \cellcolor{magenta!10}54.48$\pm{6.20}$   & \cellcolor{magenta!8}57.03$\pm{3.50}$   & \cellcolor{magenta!8}56.79$\pm{2.58}$  & \cellcolor{magenta!13}52.19$\pm{2.12}$ & \textbf{64.33}$\pm{5.92}$  &     7 / 7\\ \midrule
                                     & & 0\% (zero-shot)                    &        -          &     -        &    -         &      -       &     -        &     -       &       -      & \textbf{62.64}$\pm{9.65}$ &                                      \\
                                     &\multirow{-1}{*}{ACC $\uparrow$}& 1\%                    & \cellcolor{teal!4}66.41$\pm{2.20}$  & \cellcolor{teal!4}\textbf{67.10}$\pm{10.90}$ & \cellcolor{teal!1}63.20$\pm{3.32}$  & \cellcolor{teal!4}66.56$\pm{2.19}$  & \cellcolor{magenta!0}62.34$\pm{2.81}$     & \cellcolor{teal!4}66.14$\pm{2.13}$ & \cellcolor{magenta!2}60.18$\pm{2.54}$ & 62.75$\pm{4.92}$  &     2 / 7\\
\multirow{-1}{*}[-0.4\baselineskip]{SanDiego}           & & 10\%                   & \cellcolor{teal!5}67.17$\pm{2.90}$   & \cellcolor{teal!4}66.61$\pm{10.01}$ & \cellcolor{teal!2}65.00$\pm{2.21}$  & \cellcolor{teal!4}66.69$\pm{2.39}$  & \cellcolor{teal!1}63.17$\pm{0.83}$  & \cellcolor{teal!4}66.34$\pm{2.20}$ & \cellcolor{magenta!2}61.06$\pm{2.85}$ & \textbf{67.30}$\pm{5.32}$   &     1 / 7\\ \cmidrule(lr){2-12}
& & 0\% (zero-shot)                    &        -          &     -        &    -         &      -       &     -        &     -       &       -      & \textbf{63.61}$\pm{11.13}$ &                                      \\
                                     &\multirow{-1}{*}{AUC $\uparrow$}& 1\%                    & \cellcolor{magenta!4}59.21$\pm{4.15}$  & \cellcolor{magenta!11}52.75$\pm{22.95}$ & \cellcolor{magenta!12}51.42$\pm{3.61}$  & \cellcolor{magenta!6}57.49$\pm{4.83}$  & \cellcolor{magenta!7}56.17$\pm{5.00}$     & \cellcolor{magenta!2}61.55$\pm{11.61}$ & \cellcolor{magenta!9}54.66$\pm{6.61}$ & \textbf{63.82}$\pm{5.67}$  &     7 / 7\\
& & 10\%                   & \cellcolor{magenta!1}62.30$\pm{8.12}$   & \cellcolor{magenta!9}54.49$\pm{22.51}$ & \cellcolor{magenta!12}51.53$\pm{3.62}$  & \cellcolor{magenta!2}61.59$\pm{4.23}$  & \cellcolor{magenta!6}57.91$\pm{2.96}$  & \cellcolor{magenta!2}62.05$\pm{11.42}$ & \cellcolor{magenta!9}54.39$\pm{6.52}$ & \textbf{65.99}$\pm{4.80}$   &     7 / 7\\ \midrule
                                     & & 0\% (zero-shot)                    &        -          &     -        &    -         &      -       &     -        &     -       &       -    & \textbf{59.13}$\pm{3.37}$  &                                      \\
                                     &\multirow{-1}{*}{ACC $\uparrow$}& 1\%                    & \cellcolor{magenta!0}59.32$\pm{3.21}$  & \cellcolor{teal!8}67.27$\pm{4.85}$  & \cellcolor{magenta!4}55.37$\pm{1.22}$  & \cellcolor{magenta!1}58.71$\pm{4.62}$  & \cellcolor{magenta!4}55.80$\pm{0.73}$  & \cellcolor{magenta!5}54.60$\pm{5.78}$     & \cellcolor{teal!1}60.14$\pm{2.34}$ & \textbf{68.91}$\pm{3.48}$  &     4 / 7\\
\multirow{-1}{*}[-0.4\baselineskip]{ADHD}      & & 10\%                  & \cellcolor{teal!4}63.07$\pm{3.21}$  & \cellcolor{teal!7}65.95$\pm{5.75}$  & \cellcolor{magenta!3}56.10$\pm{1.57}$  & \cellcolor{teal!12}71.13$\pm{6.52}$  & \cellcolor{teal!2}61.02$\pm{3.79}$  & \cellcolor{teal!3}62.14$\pm{2.69}$   & \cellcolor{teal!2}60.96$\pm{3.29}$  & \textbf{74.24}$\pm{3.87}$  &     1 / 7\\ \cmidrule(lr){2-12}
& & 0\% (zero-shot)                    &        -          &     -        &    -         &      -       &     -        &     -       &       -    & \textbf{59.44}$\pm{3.83}$  &                                      \\
                                     &\multirow{-1}{*}{AUC $\uparrow$}& 1\%                    & \cellcolor{magenta!3}56.23$\pm{9.65}$  & \cellcolor{teal!9}68.11$\pm{7.39}$  & \cellcolor{magenta!7}52.06$\pm{3.82}$  & \cellcolor{magenta!0}58.95$\pm{8.94}$  & \cellcolor{magenta!3}56.56$\pm{2.16}$  & \cellcolor{magenta!6}52.81$\pm{3.00}$     & \cellcolor{magenta!2}56.72$\pm{5.23}$ & \textbf{72.74}$\pm{5.11}$  &     6 / 7\\
& & 10\%                  & \cellcolor{teal!5}63.68$\pm{4.97}$  & \cellcolor{teal!10}68.95$\pm{7.95}$  & \cellcolor{magenta!4}55.19$\pm{2.31}$  & \cellcolor{teal!15}73.77$\pm{9.29}$  & \cellcolor{teal!2}60.94$\pm{6.29}$  & \cellcolor{teal!3}61.85$\pm{4.50}$   & \cellcolor{teal!1}59.63$\pm{5.30}$  & \textbf{79.25}$\pm{4.14}$  &     1 / 7\\ \midrule

                                     & & 0\% (zero-shot)                    &      -          &     -        &    -         &      -       &     -        &     -       &       -     & \textbf{62.23}$\pm{6.49}$  &                                      \\
                                     &\multirow{-1}{*}{ACC $\uparrow$}& 1\%                    & \cellcolor{magenta!2}60.61$\pm{5.82}$  & \cellcolor{magenta!1}60.85$\pm{7.57}$ & \cellcolor{magenta!6}55.94$\pm{2.75}$   & \cellcolor{teal!4}\textbf{65.99}$\pm{9.66}$  & \cellcolor{magenta!5}57.62$\pm{5.21}$ & \cellcolor{teal!1}63.16$\pm{4.35}$     & \cellcolor{magenta!9}53.52$\pm{4.95}$   & 61.91$\pm{5.68}$   &     5 / 7\\
\multirow{-1}{*}[-0.4\baselineskip]{Schizo}      & & 10\%                   & \cellcolor{teal!7}\textbf{69.55}$\pm{7.08}$  & \cellcolor{teal!1}63.20$\pm{10.19}$  & \cellcolor{magenta!5}56.92$\pm{3.70}$  & \cellcolor{teal!5}67.61$\pm{8.62}$ & \cellcolor{magenta!4}58.14$\pm{4.57}$  & \cellcolor{teal!2}63.93$\pm{4.08}$  & \cellcolor{magenta!13}49.02$\pm{7.14}$ & 66.62$\pm{8.96}$ &     3 / 7\\ \cmidrule(lr){2-12}
& & 0\% (zero-shot)                    &      -          &     -        &    -         &      -       &     -        &     -       &       -     & \textbf{73.53}$\pm{8.35}$  &                                      \\
                                     &\multirow{-1}{*}{AUC $\uparrow$}& 1\%                    & \cellcolor{magenta!6}67.83$\pm{6.86}$  & \cellcolor{magenta!14}59.66$\pm{13.77}$ & \cellcolor{magenta!17}56.10$\pm{6.09}$   & \cellcolor{magenta!7}66.80$\pm{11.39}$  & \cellcolor{magenta!17}56.26$\pm{10.36}$ & \cellcolor{magenta!6}67.24$\pm{8.00}$     & \cellcolor{magenta!22}51.70$\pm{4.70}$   & \textbf{69.43}$\pm{9.80}$   &     7 / 7\\
                                     & & 10\%                   & \cellcolor{magenta!1}72.34$\pm{9.11}$  & \cellcolor{magenta!11}62.90$\pm{16.42}$  & \cellcolor{magenta!19}54.81$\pm{7.65}$  & \cellcolor{magenta!4}69.43$\pm{11.93}$ & \cellcolor{magenta!17}56.99$\pm{6.35}$  & \cellcolor{magenta!6}67.33$\pm{6.21}$  & \cellcolor{magenta!25}48.31$\pm{8.48}$ & \textbf{74.19}$\pm{13.74}$ &     7 / 7\\ 
                                     
 \midrule

                                     & & 0\% (zero-shot)                    &       -          &     -        &    -         &      -       &     -        &     -       &       -     & \textbf{73.20}$\pm{12.71}$   &                                      \\
                                     &\multirow{-1}{*}{ACC $\uparrow$}& 1\%                    & \cellcolor{magenta!2}71.09$\pm{9.28}$  & \cellcolor{teal!7}\textbf{79.75}$\pm{13.51}$ & \cellcolor{magenta!11}61.86$\pm{10.94}$  & \cellcolor{magenta!4}68.96$\pm{6.05}$ & \cellcolor{magenta!19}53.72$\pm{5.58}$  & \cellcolor{magenta!11}62.54$\pm{9.05}$  & \cellcolor{magenta!18}55.26$\pm{4.75}$ & 68.29$\pm{15.83}$ &     6 / 7\\
\multirow{-1}{*}[-0.4\baselineskip]{APAVA}              & & 10\%                   & \cellcolor{teal!1}74.50$\pm{16.39}$  & \cellcolor{teal!5}78.57$\pm{14.37}$ & \cellcolor{magenta!10}62.78$\pm{9.58}$  & \cellcolor{magenta!4}69.69$\pm{11.59}$ & \cellcolor{magenta!13}60.07$\pm{7.40}$ & \cellcolor{magenta!9}64.45$\pm{9.82}$ & \cellcolor{magenta!17}56.55$\pm{5.10}$ & \textbf{78.87}$\pm{15.56}$  &     5 / 7\\ \cmidrule(lr){2-12}
& & 0\% (zero-shot)                    &       -          &     -        &    -         &      -       &     -        &     -       &       -     & \textbf{82.36}$\pm{6.60}$   &                                      \\
                                     &\multirow{-1}{*}{AUC $\uparrow$}& 1\%                    & \cellcolor{teal!0}82.76$\pm{6.31}$  & \cellcolor{teal!6}\textbf{88.56}$\pm{12.02}$ & \cellcolor{magenta!24}58.58$\pm{3.24}$  & \cellcolor{magenta!13}69.29$\pm{12.93}$ & \cellcolor{magenta!30}52.44$\pm{9.26}$  & \cellcolor{magenta!14}67.97$\pm{14.80}$  & \cellcolor{magenta!29}53.59$\pm{5.27}$ & 73.39$\pm{17.66}$ &     5 / 7\\
& & 10\%                   & \cellcolor{teal!7}88.93$\pm{4.96}$  & \cellcolor{teal!6}88.24$\pm{13.39}$ & \cellcolor{magenta!20}62.24$\pm{7.79}$  & \cellcolor{magenta!7}75.55$\pm{15.15}$ & \cellcolor{magenta!21}61.23$\pm{12.09}$ & \cellcolor{magenta!14}68.09$\pm{15.04}$ & \cellcolor{magenta!27}55.11$\pm{4.45}$ & \textbf{91.22}$\pm{5.34}$  &     5 / 7\\ \midrule

                                     & & 0\% (zero-shot)                    &     -          &     -        &    -         &      -       &     -        &     -       &       -    & \textbf{80.05}$\pm{8.67}$ &                                      \\
                                     &\multirow{-1}{*}{ACC $\uparrow$}& 1\%                    & \cellcolor{magenta!5}74.96$\pm{14.82}$ & \cellcolor{teal!4}84.12$\pm{6.44}$ & \cellcolor{teal!6}86.05$\pm{2.17}$  & \cellcolor{teal!8}87.57$\pm{3.27}$  & \cellcolor{teal!3}83.46$\pm{7.48}$  & \cellcolor{teal!7}87.02$\pm{2.56}$   & \cellcolor{magenta!0}79.58$\pm{5.14}$ & \textbf{88.41}$\pm{3.99}$ &     2 / 7\\
\multirow{-1}{*}[-0.4\baselineskip]{ADFSU}              & & 10\%                   & \cellcolor{teal!1}80.62$\pm{9.23}$ & \cellcolor{teal!7}86.98$\pm{2.63}$ & \cellcolor{teal!7}86.88$\pm{2.50}$  & \cellcolor{teal!10}\textbf{90.26}$\pm{2.32}$ & \cellcolor{teal!10}90.00$\pm{3.51}$  & \cellcolor{teal!7}87.02$\pm{2.56}$  & \cellcolor{teal!3}83.29$\pm{3.08}$ & 87.99$\pm{3.60}$   &     0 / 7\\ \cmidrule(lr){2-12}
& & 0\% (zero-shot)                    &     -          &     -        &    -         &      -       &     -        &     -       &       -    & \textbf{71.79}$\pm{15.45}$ &                                      \\
                                     &\multirow{-1}{*}{AUC $\uparrow$}& 1\%                    & \cellcolor{teal!0}72.01$\pm{18.47}$ & \cellcolor{magenta!20}51.46$\pm{20.76}$ & \cellcolor{magenta!9}62.58$\pm{2.92}$  & \cellcolor{magenta!11}60.80$\pm{14.58}$  & \cellcolor{magenta!2}70.06$\pm{9.25}$  & \cellcolor{magenta!18}53.57$\pm{7.30}$   & \cellcolor{magenta!21}50.87$\pm{5.28}$ & \textbf{78.96}$\pm{11.29}$ &     6 / 7\\
& & 10\%                   & \cellcolor{teal!4}76.13$\pm{11.92}$ & \cellcolor{magenta!8}64.27$\pm{14.87}$ & \cellcolor{magenta!8}64.18$\pm{5.62}$  & \cellcolor{teal!12}84.19$\pm{13.04}$ & \cellcolor{teal!12}83.54$\pm{8.19}$  & \cellcolor{magenta!20}51.43$\pm{6.55}$  & \cellcolor{magenta!21}50.99$\pm{4.13}$ & \textbf{86.16}$\pm{9.40}$   &     4 / 7\\ \midrule
                                     & & 0\% (zero-shot)                    &       -          &     -        &    -         &      -       &     -        &     -       &       -     & \textbf{67.80}$\pm{6.32}$     &                                      \\
                                     &\multirow{-1}{*}{ACC $\uparrow$}& 1\%                    & \cellcolor{magenta!1}66.46$\pm{12.22}$  & \cellcolor{teal!0}68.17$\pm{11.09}$  & \cellcolor{magenta!3}64.39$\pm{9.05}$  & \cellcolor{magenta!2}65.63$\pm{10.93}$   & \cellcolor{magenta!3}64.97$\pm{9.28}$  & \cellcolor{magenta!1}66.52$\pm{12.31}$  & \cellcolor{magenta!7}61.28$\pm{5.83}$  & \textbf{68.82}$\pm{7.96}$  &     6 / 7\\
\multirow{-1}{*}[-0.4\baselineskip]{IEDS}               & & 10\%                   & \cellcolor{magenta!1}66.46$\pm{12.29}$  & \cellcolor{teal!2}69.30$\pm{9.37}$   & \cellcolor{magenta!2}66.00$\pm{10.95}$  & \cellcolor{magenta!1}66.62$\pm{12.29}$  & \cellcolor{magenta!2}65.96$\pm{9.65}$  & \cellcolor{magenta!1}67.03$\pm{12.17}$  & \cellcolor{magenta!4}63.86$\pm{8.47}$ & \textbf{69.94}$\pm{9.50}$  &     6 / 7\\ \cmidrule(lr){2-12}
& & 0\% (zero-shot)                    &       -          &     -        &    -         &      -       &     -        &     -       &       -     & \textbf{61.00}$\pm{4.98}$     &                                      \\
                                     &\multirow{-1}{*}{AUC $\uparrow$}& 1\%                    & \cellcolor{magenta!12}49.44$\pm{1.38}$  & \cellcolor{magenta!1}\textbf{59.76}$\pm{4.23}$  & \cellcolor{magenta!9}51.81$\pm{1.49}$  & \cellcolor{magenta!9}52.48$\pm{2.20}$   & \cellcolor{magenta!7}54.46$\pm{2.44}$  & \cellcolor{magenta!2}58.78$\pm{6.33}$  & \cellcolor{magenta!9}51.51$\pm{1.10}$  & 59.55$\pm{2.58}$  &     7 / 7\\
& & 10\%                   & \cellcolor{magenta!11}49.87$\pm{2.16}$  & \cellcolor{magenta!1}60.13$\pm{1.80}$   & \cellcolor{magenta!8}53.12$\pm{0.68}$  & \cellcolor{magenta!17}44.48$\pm{5.89}$  & \cellcolor{magenta!6}55.14$\pm{2.73}$  & \cellcolor{magenta!2}58.93$\pm{5.53}$  & \cellcolor{magenta!7}54.15$\pm{1.54}$ & \textbf{61.57}$\pm{4.68}$  &     7 / 7\\ \midrule

                                     & & 0\% (zero-shot)                    &      -          &     -        &    -         &      -       &     -        &     -       &       -      & \textbf{77.26}$\pm{6.76}$  &                                      \\
                                     &\multirow{-1}{*}{ACC $\uparrow$}& 1\%                    & \cellcolor{magenta!4}72.79$\pm{27.54}$  & \cellcolor{magenta!5}72.41$\pm{27.52}$ & \cellcolor{magenta!2}75.14$\pm{21.73}$   & \cellcolor{magenta!50}27.76$\pm{27.42}$  & \cellcolor{magenta!4}72.83$\pm{26.63}$   & \cellcolor{magenta!5}72.41$\pm{27.52}$  & \cellcolor{magenta!3}74.70$\pm{22.64}$ & \textbf{88.35}$\pm{3.25}$  &     7 / 7\\
\multirow{-1}{*}[-0.4\baselineskip]{Mayo}               & & 10\%                   & \cellcolor{teal!0}77.73$\pm{21.05}$   & \cellcolor{magenta!5}72.41$\pm{27.52}$ & \cellcolor{magenta!5}72.41$\pm{27.52}$  & \cellcolor{teal!5}82.38$\pm{8.64}$  & \cellcolor{magenta!4}72.92$\pm{27.30}$  & \cellcolor{magenta!4}72.77$\pm{27.52}$   & \cellcolor{magenta!5}72.41$\pm{27.52}$ & \textbf{87.50}$\pm{7.22}$  &     5 / 7\\ \cmidrule(lr){2-12}
& & 0\% (zero-shot)                    &      -          &     -        &    -         &      -       &     -        &     -       &       -      & \textbf{88.31}$\pm{4.24}$  &                                      \\
                                     &\multirow{-1}{*}{AUC $\uparrow$}& 1\%                    & \cellcolor{magenta!19}69.14$\pm{6.19}$  & \cellcolor{magenta!58}30.08$\pm{23.12}$ & \cellcolor{magenta!41}47.71$\pm{3.80}$   & \cellcolor{magenta!52}36.50$\pm{10.56}$  & \cellcolor{magenta!36}51.88$\pm{3.90}$   & \cellcolor{magenta!26}62.55$\pm{6.48}$  & \cellcolor{magenta!36}51.82$\pm{2.67}$ & \textbf{89.64}$\pm{2.57}$  &     7 / 7\\
& & 10\%                   & \cellcolor{magenta!12}75.91$\pm{7.80}$   & \cellcolor{magenta!11}76.95$\pm{14.68}$ & \cellcolor{magenta!33}55.58$\pm{4.07}$  & \cellcolor{magenta!24}64.55$\pm{9.22}$  & \cellcolor{magenta!26}62.73$\pm{8.67}$  & \cellcolor{magenta!14}74.33$\pm{3.50}$   & \cellcolor{magenta!36}52.27$\pm{2.36}$ & \textbf{90.64}$\pm{1.91}$  &     7 / 7\\ \midrule

                                     & & 0\% (zero-shot)                    &      -          &     -        &    -         &      -       &     -        &     -       &       -    & \textbf{57.47}$\pm{0.00}$      &                                      \\
                                     &\multirow{-1}{*}{ACC $\uparrow$}& 1\%                    & \cellcolor{magenta!6}51.27$\pm{0.64}$  & \cellcolor{magenta!0}57.28$\pm{4.36}$  & \cellcolor{magenta!2}55.74$\pm{5.16}$  & \cellcolor{teal!7}64.69$\pm{3.52}$  & \cellcolor{magenta!1}56.00$\pm{1.08}$  & \cellcolor{magenta!4}53.32$\pm{3.22}$ & \cellcolor{magenta!2}55.22$\pm{1.13}$ & \textbf{66.39}$\pm{0.53}$     &     6 / 7\\
\multirow{-1}{*}[-0.4\baselineskip]{NMT}                & & 10\%                   & \cellcolor{magenta!4}53.36$\pm{1.36}$  & \cellcolor{teal!1}58.41$\pm{2.04}$  & \cellcolor{magenta!5}52.72$\pm{1.77}$  & \cellcolor{teal!9}\textbf{66.87}$\pm{1.01}$  & \cellcolor{teal!1}58.13$\pm{1.28}$   & \cellcolor{magenta!5}52.68$\pm{0.24}$  & \cellcolor{magenta!2}55.78$\pm{0.39}$  & 63.84$\pm{0.47}$     &     4 / 7\\ \cmidrule(lr){2-12}
& & 0\% (zero-shot)                    &      -          &     -        &    -         &      -       &     -        &     -       &       -    & \textbf{75.10}$\pm{0.00}$      &                                      \\
                                     &\multirow{-1}{*}{AUC $\uparrow$}& 1\%                    & \cellcolor{teal!1}76.55$\pm{0.86}$  & \cellcolor{magenta!4}71.24$\pm{1.81}$  & \cellcolor{magenta!8}66.94$\pm{3.64}$  & \cellcolor{magenta!2}72.98$\pm{0.84}$  & \cellcolor{magenta!9}65.95$\pm{0.79}$  & \cellcolor{magenta!16}59.06$\pm{12.16}$ & \cellcolor{magenta!16}58.61$\pm{1.72}$ & \textbf{76.98}$\pm{0.15}$     &     6 / 7\\
& & 10\%                   & \cellcolor{teal!2}76.85$\pm{0.41}$  & \cellcolor{magenta!4}71.54$\pm{1.05}$  & \cellcolor{magenta!6}69.31$\pm{1.77}$  & \cellcolor{magenta!4}71.12$\pm{0.59}$  & \cellcolor{magenta!4}70.80$\pm{0.55}$   & \cellcolor{magenta!0}74.91$\pm{0.36}$  & \cellcolor{magenta!13}62.40$\pm{2.14}$  & \textbf{79.79}$\pm{0.32}$     &     6 / 7\\  \midrule

                                     & & 0\% (zero-shot)                    &       -          &     -        &    -         &      -       &     -        &     -       &       -     & \textbf{61.45}$\pm{0.00}$     &                                      \\
                                     &\multirow{-1}{*}{ACC $\uparrow$}& 1\%                    & \cellcolor{teal!1}62.61$\pm{7.12}$ & \cellcolor{teal!5}\textbf{66.08}$\pm{5.92}$   & \cellcolor{magenta!9}52.28$\pm{1.55}$  & \cellcolor{magenta!7}54.36$\pm{4.61}$  & \cellcolor{magenta!6}55.03$\pm{5.41}$  & \cellcolor{magenta!1}60.41$\pm{2.69}$  & \cellcolor{magenta!10}51.10$\pm{1.99}$  & 65.15$\pm{0.97}$  &     5 / 7\\
\multirow{-1}{*}[-0.4\baselineskip]{SEE}                & & 10\%                   & \cellcolor{teal!9}70.77$\pm{0.78}$  & \cellcolor{teal!7}68.74$\pm{3.86}$  & \cellcolor{magenta!2}59.84$\pm{5.17}$  & \cellcolor{teal!1}62.41$\pm{2.17}$  & \cellcolor{magenta!1}60.28$\pm{1.11}$  & \cellcolor{teal!2}63.87$\pm{4.04}$  & \cellcolor{magenta!10}51.50$\pm{2.29}$ & \textbf{72.66}$\pm{0.72}$  &     3 / 7\\ \cmidrule(lr){2-12}
& & 0\% (zero-shot)                    &       -          &     -        &    -         &      -       &     -        &     -       &       -     & \textbf{64.24}$\pm{0.00}$     &                                      \\
                                     &\multirow{-1}{*}{AUC $\uparrow$}& 1\%                    & \cellcolor{teal!4}68.09$\pm{10.35}$ & \cellcolor{teal!11}\textbf{74.92}$\pm{7.70}$   & \cellcolor{magenta!11}53.63$\pm{1.96}$  & \cellcolor{magenta!0}64.06$\pm{7.97}$  & \cellcolor{magenta!7}56.89$\pm{8.01}$  & \cellcolor{teal!3}66.84$\pm{3.64}$  & \cellcolor{magenta!13}51.10$\pm{2.15}$  & 72.19$\pm{1.82}$  &     4 / 7\\
& & 10\%                   & \cellcolor{teal!13}76.83$\pm{1.06}$  & \cellcolor{teal!15}79.14$\pm{2.32}$  & \cellcolor{magenta!1}63.38$\pm{6.87}$  & \cellcolor{teal!7}71.26$\pm{0.89}$  & \cellcolor{magenta!0}63.77$\pm{1.83}$  & \cellcolor{teal!4}68.52$\pm{5.83}$  & \cellcolor{magenta!13}51.61$\pm{2.68}$ & \textbf{79.57}$\pm{0.63}$  &     3 / 7\\ \midrule
                                     & & 0\% (zero-shot)                    &      -          &     -        &    -         &      -       &     -        &     -       &       -    & \textbf{67.41}    &                                      \\
                                     &\multirow{-1}{*}{ACC $\uparrow$}& 1\%                    & \cellcolor{magenta!5}62.47    & \cellcolor{magenta!4}63.56    & \cellcolor{magenta!8}59.24    & \cellcolor{magenta!10}57.58    & \cellcolor{magenta!10}57.43    & \cellcolor{magenta!8}59.56    & \cellcolor{magenta!12}55.06   & \textbf{70.74}    &     7 / 7\\
\multirow{-1}{*}[-0.4\baselineskip]{\textbf{Ave}}                & & 10\%                   & \cellcolor{teal!0}67.69    & \cellcolor{magenta!0}67.15    & \cellcolor{magenta!4}63.15     & \cellcolor{magenta!3}64.61    & \cellcolor{magenta!5}62.52     & \cellcolor{magenta!4}63.27    & \cellcolor{magenta!11}56.00   & \textbf{73.78}    &     6 / 7\\ \cmidrule(lr){2-12}
& & 0\% (zero-shot)                    &      -          &     -        &    -         &      -       &     -        &     -       &       -    & \textbf{74.33}    &                                      \\
                                     &\multirow{-1}{*}{AUC $\uparrow$}& 1\%                    & \cellcolor{magenta!8}68.36    & \cellcolor{magenta!11}63.02    & \cellcolor{magenta!17}57.54    & \cellcolor{magenta!15}59.43    & \cellcolor{magenta!16}58.15    & \cellcolor{magenta!13}61.36    & \cellcolor{magenta!21}52.92   & \textbf{75.59}    &     7 / 7\\
& & 10\%                   & \cellcolor{magenta!1}73.23    & \cellcolor{magenta!3}71.59    & \cellcolor{magenta!12}62.09     & \cellcolor{magenta!8}68.36    & \cellcolor{magenta!9}65.43     & \cellcolor{magenta!8}66.75    & \cellcolor{magenta!20}54.19   & \textbf{80.19}    &     7 / 7\\ \bottomrule
\end{tabular}%
}
\end{table}

\newpage
\section{Reasoning Analysis}
\label{app_reasoning_analysis}

\subsection{Per-dataset Reasoning Analysis Result}
\begin{table}[htb]
\caption{\textbf{The results of different reasoning settings on 12 downstream EEG/iEEG datasets.}}
\label{app_result_reasoning_analysis}
\resizebox{\textwidth}{!}{%
\begin{tabular}{@{}ccccccccc@{}}
\toprule
\textbf{Variants}     & \multicolumn{2}{c}{\textbf{Stage-\RNum{1} Direct}} & \multicolumn{2}{c}{\textbf{Stage-\RNum{2} Direct}} & \multicolumn{2}{c}{\textbf{Stage-\RNum{2} Intervention}} & \multicolumn{2}{c}{\textbf{Stage-\RNum{2} Reasoning}} \\ \midrule
Datasets     & ACC & AUC & ACC & AUC & ACC & AUC & ACC & AUC \\ \midrule
ISRUC        & \textbf{83.30}$\pm\scriptstyle{1.19}$ & \cellcolor{gray!15}94.92$\pm\scriptstyle{0.49}$ & 81.80$\pm\scriptstyle{0.83}$ & \cellcolor{gray!15}94.76$\pm\scriptstyle{0.46}$ & 75.63$\pm\scriptstyle{0.74}$ & \cellcolor{gray!15}90.91$\pm\scriptstyle{0.45}$ & 82.77$\pm\scriptstyle{1.38}$ & \cellcolor{gray!15}\textbf{95.19$\pm\scriptstyle{0.62}$} \\
Dreams       & 73.27$\pm\scriptstyle{2.96}$ & \cellcolor{gray!15}90.10$\pm\scriptstyle{0.90}$ & 75.77$\pm\scriptstyle{2.20}$ & \cellcolor{gray!15}91.38$\pm\scriptstyle{0.87}$ & 67.83$\pm\scriptstyle{1.97}$ & \cellcolor{gray!15}85.85$\pm\scriptstyle{1.22}$ & \textbf{77.66}$\pm\scriptstyle{2.51}$ & \cellcolor{gray!15}\textbf{92.49$\pm\scriptstyle{1.07}$} \\
Siena        & 46.69$\pm\scriptstyle{11.25}$ & \cellcolor{gray!15}60.13$\pm\scriptstyle{9.48}$ & 47.45$\pm\scriptstyle{12.41}$ & \cellcolor{gray!15}58.99$\pm\scriptstyle{8.01}$ & \textbf{47.77}$\pm\scriptstyle{12.91}$ & \cellcolor{gray!15}56.58$\pm\scriptstyle{6.81}$ & 47.31$\pm\scriptstyle{16.20}$ & \cellcolor{gray!15}\textbf{64.85$\pm\scriptstyle{9.68}$} \\
SanDiego     & 50.40$\pm\scriptstyle{10.69}$ & \cellcolor{gray!15}57.64$\pm\scriptstyle{11.58}$ & 56.74$\pm\scriptstyle{5.80}$ & \cellcolor{gray!15}59.87$\pm\scriptstyle{6.18}$ & 58.66$\pm\scriptstyle{4.61}$ & \cellcolor{gray!15}58.67$\pm\scriptstyle{6.10}$ & \textbf{62.64}$\pm\scriptstyle{9.65}$ & \cellcolor{gray!15}\textbf{63.61$\pm\scriptstyle{11.13}$} \\
ADHD         & 47.95$\pm\scriptstyle{5.17}$ & \cellcolor{gray!15}45.32$\pm\scriptstyle{6.85}$ & 47.72$\pm\scriptstyle{4.68}$ & \cellcolor{gray!15}44.13$\pm\scriptstyle{7.11}$ & 50.92$\pm\scriptstyle{3.40}$ & \cellcolor{gray!15}50.68$\pm\scriptstyle{6.08}$ & \textbf{59.13}$\pm\scriptstyle{3.37}$ & \cellcolor{gray!15}\textbf{59.44$\pm\scriptstyle{3.83}$} \\
Schizo       & 47.07$\pm\scriptstyle{8.64}$ & \cellcolor{gray!15}52.86$\pm\scriptstyle{16.70}$ & 56.70$\pm\scriptstyle{8.93}$ & \cellcolor{gray!15}62.51$\pm\scriptstyle{11.33}$ & 59.80$\pm\scriptstyle{8.85}$ & \cellcolor{gray!15}62.48$\pm\scriptstyle{12.85}$ & \textbf{62.23}$\pm\scriptstyle{6.49}$ & \cellcolor{gray!15}\textbf{73.53$\pm\scriptstyle{8.35}$} \\
APAVA        & 45.12$\pm\scriptstyle{11.29}$ & \cellcolor{gray!15}35.43$\pm\scriptstyle{14.23}$ & 63.78$\pm\scriptstyle{16.69}$ & \cellcolor{gray!15}68.54$\pm\scriptstyle{19.76}$ & 65.64$\pm\scriptstyle{16.89}$ & \cellcolor{gray!15}69.13$\pm\scriptstyle{22.41}$ & \textbf{73.20}$\pm\scriptstyle{12.71}$ & \cellcolor{gray!15}\textbf{82.36$\pm\scriptstyle{6.60}$} \\
ADFSU        & 57.58$\pm\scriptstyle{10.67}$ & \cellcolor{gray!15}55.21$\pm\scriptstyle{11.67}$ & 71.10$\pm\scriptstyle{12.06}$ & \cellcolor{gray!15}69.80$\pm\scriptstyle{17.33}$ & 77.11$\pm\scriptstyle{11.60}$ & \cellcolor{gray!15}69.74$\pm\scriptstyle{17.46}$ & \textbf{80.05}$\pm\scriptstyle{8.67}$ & \cellcolor{gray!15}\textbf{71.79$\pm\scriptstyle{15.45}$} \\
IEDS         & 63.09$\pm\scriptstyle{5.91}$ & \cellcolor{gray!15}58.67$\pm\scriptstyle{3.35}$ & 63.66$\pm\scriptstyle{5.50}$ & \cellcolor{gray!15}59.44$\pm\scriptstyle{2.87}$ & 57.15$\pm\scriptstyle{4.05}$ & \cellcolor{gray!15}59.10$\pm\scriptstyle{2.27}$ & \textbf{67.80}$\pm\scriptstyle{6.32}$ & \cellcolor{gray!15}\textbf{61.00$\pm\scriptstyle{4.98}$} \\
Mayo         & 62.27$\pm\scriptstyle{13.18}$ & \cellcolor{gray!15}80.94$\pm\scriptstyle{8.73}$ & 71.23$\pm\scriptstyle{9.72}$ & \cellcolor{gray!15}87.88$\pm\scriptstyle{4.61}$ & 59.61$\pm\scriptstyle{17.55}$ & \cellcolor{gray!15}87.10$\pm\scriptstyle{4.65}$ & \textbf{77.26}$\pm\scriptstyle{6.76}$ & \cellcolor{gray!15}\textbf{88.31$\pm\scriptstyle{4.24}$} \\
NMT          & 52.77$\pm\scriptstyle{0.00}$ & \cellcolor{gray!15}62.25$\pm\scriptstyle{0.00}$ & 57.50$\pm\scriptstyle{0.00}$ & \cellcolor{gray!15}71.47$\pm\scriptstyle{0.00}$ & \textbf{58.60}$\pm\scriptstyle{0.13}$ & \cellcolor{gray!15}70.95$\pm\scriptstyle{0.04}$ & 57.47$\pm\scriptstyle{0.00}$ & \cellcolor{gray!15}\textbf{75.10$\pm\scriptstyle{0.00}$} \\
SEE          & \textbf{61.56}$\pm\scriptstyle{0.22}$ & \cellcolor{gray!15}64.63$\pm\scriptstyle{0.06}$ & 61.50$\pm\scriptstyle{0.00}$ & \cellcolor{gray!15}\textbf{65.98$\pm\scriptstyle{0.00}$} & 49.85$\pm\scriptstyle{0.00}$ & \cellcolor{gray!15}54.36$\pm\scriptstyle{0.00}$ & 61.45$\pm\scriptstyle{0.00}$ & \cellcolor{gray!15}64.24$\pm\scriptstyle{0.00}$ \\ \midrule
Ave          & 57.59 & 63.18 & 62.91 & 69.56 & 60.71 & 67.96 & \textbf{67.41} & \textbf{74.33} \\ \bottomrule
\end{tabular}%
}
\end{table}

\subsection{Machine Learning Classification using Features}

\begin{table}[htb]
\centering
\caption{\textbf{AUC performance of classical machine learning models on downstream datasets.}}
\label{tab:ml_baselines}
\resizebox{\textwidth}{!}{%
\begin{tabular}{ccccccc}
\toprule
\textbf{Dataset} & \textbf{Logistic Regression} & \textbf{KNN} & \textbf{MLP} & \textbf{SVM} & \textbf{Random Forest} & \textbf{XGBoost} \\
\midrule
ISRUC  & 91.74$\pm\scriptstyle{0.94}$ & 85.29$\pm\scriptstyle{1.48}$ & 90.46$\pm\scriptstyle{0.60}$ & 92.38$\pm\scriptstyle{1.05}$ & 91.90$\pm\scriptstyle{1.24}$ & 92.93$\pm\scriptstyle{1.03}$ \\
Dreams & 91.16$\pm\scriptstyle{1.35}$ & 83.67$\pm\scriptstyle{1.29}$ & 88.94$\pm\scriptstyle{1.72}$ & 91.15$\pm\scriptstyle{0.81}$ & 91.15$\pm\scriptstyle{1.05}$ & 91.59$\pm\scriptstyle{0.40}$ \\
Siena  & 59.88$\pm\scriptstyle{4.11}$ & 53.43$\pm\scriptstyle{2.28}$ & 56.85$\pm\scriptstyle{3.89}$ & 59.11$\pm\scriptstyle{4.37}$ & 58.31$\pm\scriptstyle{5.54}$ & 59.32$\pm\scriptstyle{6.15}$ \\
San\allowbreak Diego & 60.92$\pm\scriptstyle{6.08}$ & 63.00$\pm\scriptstyle{4.90}$ & 64.58$\pm\scriptstyle{6.84}$ & 64.84$\pm\scriptstyle{1.85}$ & 63.26$\pm\scriptstyle{5.81}$ & 67.57$\pm\scriptstyle{3.81}$ \\
ADHD   & 79.11$\pm\scriptstyle{10.04}$ & 71.07$\pm\scriptstyle{8.74}$ & 75.73$\pm\scriptstyle{9.17}$ & 77.21$\pm\scriptstyle{9.88}$ & 80.31$\pm\scriptstyle{9.09}$ & 79.57$\pm\scriptstyle{8.00}$ \\
Schizo & 67.37$\pm\scriptstyle{22.30}$ & 70.71$\pm\scriptstyle{19.75}$ & 69.09$\pm\scriptstyle{25.75}$ & 75.36$\pm\scriptstyle{17.21}$ & 74.99$\pm\scriptstyle{15.71}$ & 73.64$\pm\scriptstyle{17.63}$ \\
APAVA  & 78.77$\pm\scriptstyle{17.35}$ & 81.85$\pm\scriptstyle{7.71}$ & 85.27$\pm\scriptstyle{18.56}$ & 80.18$\pm\scriptstyle{19.48}$ & 85.31$\pm\scriptstyle{11.12}$ & 85.68$\pm\scriptstyle{10.77}$ \\
ADFSU  & 91.67$\pm\scriptstyle{4.98}$ & 76.08$\pm\scriptstyle{4.17}$ & 85.71$\pm\scriptstyle{8.24}$ & 87.89$\pm\scriptstyle{4.92}$ & 85.84$\pm\scriptstyle{6.64}$ & 88.22$\pm\scriptstyle{7.94}$ \\
IEDS   & 60.10$\pm\scriptstyle{2.97}$ & 56.42$\pm\scriptstyle{1.58}$ & 58.04$\pm\scriptstyle{2.28}$ & 63.08$\pm\scriptstyle{4.06}$ & 61.56$\pm\scriptstyle{3.52}$ & 61.39$\pm\scriptstyle{3.34}$ \\
Mayo   & 92.51$\pm\scriptstyle{2.82}$ & 84.38$\pm\scriptstyle{5.62}$ & 90.89$\pm\scriptstyle{1.85}$ & 91.93$\pm\scriptstyle{3.68}$ & 90.99$\pm\scriptstyle{4.34}$ & 92.83$\pm\scriptstyle{2.20}$ \\
NMT    & 77.25$\pm\scriptstyle{0.00}$ & 69.04$\pm\scriptstyle{0.00}$ & 77.58$\pm\scriptstyle{0.00}$ & 71.69$\pm\scriptstyle{0.00}$ & 81.70$\pm\scriptstyle{0.00}$ & 81.98$\pm\scriptstyle{0.00}$ \\
SEE    & 81.89$\pm\scriptstyle{0.00}$ & 79.13$\pm\scriptstyle{0.00}$ & 80.78$\pm\scriptstyle{0.00}$ & 78.48$\pm\scriptstyle{0.00}$ & 79.28$\pm\scriptstyle{0.00}$ & 80.07$\pm\scriptstyle{0.00}$ \\
\midrule
Ave    & 77.70 & 72.84 & 76.99 & 77.78 & 78.72 & 79.57 \\
\bottomrule
\end{tabular}
}
\end{table}

\newpage
\section{Full Fine-tuning Classification}
\label{app_full_finetuning}

\subsection{Experimental Settings}

In full fine-tuning, we unfreeze the pretrained backbone and update all trainable parameters end-to-end on each downstream dataset. A task-specific prediction head is attached and jointly optimized with the backbone, allowing the model to fully adapt its representations to the target distribution. This setting serves as an approximate upper bound under unrestricted adaptation, in contrast to linear probing where the backbone remains frozen.

\subsection{Results}

\begin{table}[htb]
\caption{\textbf{Full fine-tuning results.}}
\resizebox{\textwidth}{!}{%
\begin{tabular}{@{}cccccccccc@{}}
\toprule
\textbf{Dataset}          & \textbf{Metric} & \textbf{BrainBERT} & \textbf{BIOT} & \textbf{LaBraM} & \textbf{CBraMod} & \textbf{NeuroLM} & \textbf{ELM-MIL} & \textbf{CSBrain} & \textbf{BrainCoT} \\ \midrule
\multirow{2}{*}{ISRUC}    & ACC             & 79.18$\pm\scriptstyle{2.04}$ & 84.10$\pm\scriptstyle{3.56}$ & 79.76$\pm\scriptstyle{2.74}$ & 84.39$\pm\scriptstyle{5.17}$ & 75.27$\pm\scriptstyle{5.44}$ & \underline{84.68}$\pm\scriptstyle{3.69}$ & 77.95$\pm\scriptstyle{5.89}$ & \textbf{86.66}$\pm\scriptstyle{1.18}$ \\
                          & AUC             & 93.60$\pm\scriptstyle{0.66}$ & 95.50$\pm\scriptstyle{2.51}$ & 92.91$\pm\scriptstyle{1.90}$ & 95.85$\pm\scriptstyle{2.95}$ & 91.23$\pm\scriptstyle{3.26}$ & \underline{95.92}$\pm\scriptstyle{2.20}$ & 92.36$\pm\scriptstyle{3.21}$ & \textbf{97.00}$\pm\scriptstyle{0.32}$ \\ \midrule
\multirow{2}{*}{Dreams}   & ACC             & \underline{75.55}$\pm\scriptstyle{2.02}$ & 72.54$\pm\scriptstyle{3.20}$ & 74.56$\pm\scriptstyle{2.55}$ & 75.10$\pm\scriptstyle{2.00}$ & 40.73$\pm\scriptstyle{2.95}$ & 30.97$\pm\scriptstyle{15.37}$ & 64.67$\pm\scriptstyle{1.46}$ & \textbf{80.14}$\pm\scriptstyle{2.47}$ \\
                          & AUC             & \underline{91.58}$\pm\scriptstyle{0.90}$ & 89.99$\pm\scriptstyle{1.77}$ & 90.29$\pm\scriptstyle{1.52}$ & 90.45$\pm\scriptstyle{1.68}$ & 58.89$\pm\scriptstyle{1.35}$ & 81.45$\pm\scriptstyle{2.91}$ & 51.12$\pm\scriptstyle{0.52}$ & \textbf{94.09}$\pm\scriptstyle{0.70}$ \\ \midrule
\multirow{2}{*}{Siena}    & ACC             & 51.63$\pm\scriptstyle{14.63}$ & 49.12$\pm\scriptstyle{12.20}$ & 51.94$\pm\scriptstyle{11.46}$ & 48.13$\pm\scriptstyle{7.55}$ & \underline{52.68}$\pm\scriptstyle{12.55}$ & \textbf{56.63}$\pm\scriptstyle{11.79}$ & 48.46$\pm\scriptstyle{10.94}$ & 46.78$\pm\scriptstyle{8.32}$ \\
                          & AUC             & 55.98$\pm\scriptstyle{2.90}$ & \underline{67.39}$\pm\scriptstyle{5.47}$ & 67.29$\pm\scriptstyle{5.30}$ & 60.41$\pm\scriptstyle{4.63}$ & 66.14$\pm\scriptstyle{5.79}$ & \textbf{73.84}$\pm\scriptstyle{8.32}$ & 63.33$\pm\scriptstyle{5.41}$ & 66.08$\pm\scriptstyle{6.13}$ \\ \midrule
\multirow{2}{*}{SanDiego} & ACC             & 66.27$\pm\scriptstyle{2.14}$ & 70.78$\pm\scriptstyle{6.22}$ & 68.30$\pm\scriptstyle{3.08}$ & 69.75$\pm\scriptstyle{4.60}$ & 69.81$\pm\scriptstyle{2.91}$ & \textbf{77.35}$\pm\scriptstyle{4.54}$ & 64.29$\pm\scriptstyle{5.91}$ & \underline{75.89}$\pm\scriptstyle{7.10}$ \\
                          & AUC             & 66.30$\pm\scriptstyle{6.11}$ & 76.12$\pm\scriptstyle{5.01}$ & 64.08$\pm\scriptstyle{6.17}$ & 70.67$\pm\scriptstyle{6.36}$ & 66.71$\pm\scriptstyle{6.35}$ & \textbf{82.89}$\pm\scriptstyle{5.80}$ & 66.41$\pm\scriptstyle{6.83}$ & \underline{82.79}$\pm\scriptstyle{8.34}$ \\ \midrule
\multirow{2}{*}{ADHD}     & ACC             & 62.08$\pm\scriptstyle{4.49}$ & \underline{76.83}$\pm\scriptstyle{5.82}$ & 73.53$\pm\scriptstyle{5.12}$ & 76.73$\pm\scriptstyle{8.04}$ & 68.65$\pm\scriptstyle{2.60}$ & \textbf{83.40}$\pm\scriptstyle{3.46}$ & 73.93$\pm\scriptstyle{5.02}$ & 76.31$\pm\scriptstyle{7.57}$ \\
                          & AUC             & 66.67$\pm\scriptstyle{3.96}$ & 82.24$\pm\scriptstyle{6.38}$ & 78.17$\pm\scriptstyle{7.03}$ & \underline{84.71}$\pm\scriptstyle{6.99}$ & 73.77$\pm\scriptstyle{3.93}$ & \textbf{91.13}$\pm\scriptstyle{3.85}$ & 78.57$\pm\scriptstyle{6.81}$ & 81.83$\pm\scriptstyle{8.82}$ \\ \midrule
\multirow{2}{*}{Schizo}   & ACC             & 64.25$\pm\scriptstyle{8.24}$ & 80.50$\pm\scriptstyle{8.90}$ & 67.59$\pm\scriptstyle{7.80}$ & 80.43$\pm\scriptstyle{9.66}$ & 61.47$\pm\scriptstyle{6.00}$ & \underline{86.95}$\pm\scriptstyle{11.71}$ & 72.29$\pm\scriptstyle{8.25}$ & \textbf{88.03}$\pm\scriptstyle{10.53}$ \\
                          & AUC             & 80.30$\pm\scriptstyle{8.88}$ & 85.57$\pm\scriptstyle{8.75}$ & 71.63$\pm\scriptstyle{11.43}$ & 87.96$\pm\scriptstyle{8.63}$ & 65.82$\pm\scriptstyle{7.83}$ & \underline{93.69}$\pm\scriptstyle{10.18}$ & 73.90$\pm\scriptstyle{9.35}$ & \textbf{93.78}$\pm\scriptstyle{8.47}$ \\ \midrule
\multirow{2}{*}{APAVA}    & ACC             & \underline{81.07}$\pm\scriptstyle{9.15}$ & 73.63$\pm\scriptstyle{11.93}$ & 76.40$\pm\scriptstyle{5.76}$ & 79.69$\pm\scriptstyle{15.23}$ & 78.21$\pm\scriptstyle{4.05}$ & 80.13$\pm\scriptstyle{8.69}$ & 72.84$\pm\scriptstyle{7.60}$ & \textbf{88.63}$\pm\scriptstyle{10.40}$ \\
                          & AUC             & 93.17$\pm\scriptstyle{5.55}$ & \underline{94.73}$\pm\scriptstyle{5.43}$ & 82.82$\pm\scriptstyle{9.06}$ & 91.52$\pm\scriptstyle{11.24}$ & 84.77$\pm\scriptstyle{7.35}$ & 90.27$\pm\scriptstyle{6.38}$ & 83.18$\pm\scriptstyle{6.81}$ & \textbf{98.14}$\pm\scriptstyle{2.09}$ \\ \midrule
\multirow{2}{*}{ADFSU}    & ACC             & 89.46$\pm\scriptstyle{4.15}$ & 93.72$\pm\scriptstyle{3.44}$ & 92.16$\pm\scriptstyle{1.72}$ & 94.46$\pm\scriptstyle{2.62}$ & 94.06$\pm\scriptstyle{2.62}$ & \textbf{98.93}$\pm\scriptstyle{1.28}$ & 94.96$\pm\scriptstyle{2.16}$ & \underline{94.99}$\pm\scriptstyle{2.35}$ \\
                          & AUC             & 94.14$\pm\scriptstyle{3.26}$ & 95.63$\pm\scriptstyle{4.06}$ & 89.57$\pm\scriptstyle{2.39}$ & 94.23$\pm\scriptstyle{5.39}$ & 92.58$\pm\scriptstyle{7.92}$ & \textbf{99.71}$\pm\scriptstyle{0.55}$ & 92.08$\pm\scriptstyle{3.16}$ & \underline{97.14}$\pm\scriptstyle{3.41}$ \\ \midrule
\multirow{2}{*}{IEDS}     & ACC             & 66.45$\pm\scriptstyle{12.32}$ & \underline{70.92}$\pm\scriptstyle{6.77}$ & 67.00$\pm\scriptstyle{8.12}$ & 70.19$\pm\scriptstyle{8.10}$ & 68.49$\pm\scriptstyle{10.40}$ & \textbf{71.72}$\pm\scriptstyle{8.37}$ & 67.42$\pm\scriptstyle{10.57}$ & 67.14$\pm\scriptstyle{4.73}$ \\
                          & AUC             & 51.78$\pm\scriptstyle{0.71}$ & \underline{64.32}$\pm\scriptstyle{2.97}$ & 59.99$\pm\scriptstyle{2.11}$ & 62.19$\pm\scriptstyle{2.04}$ & 59.60$\pm\scriptstyle{2.30}$ & \textbf{65.04}$\pm\scriptstyle{3.84}$ & 57.98$\pm\scriptstyle{1.40}$ & 60.25$\pm\scriptstyle{3.80}$ \\ \midrule
\multirow{2}{*}{Mayo}     & ACC             & 72.41$\pm\scriptstyle{27.52}$ & 88.88$\pm\scriptstyle{7.76}$ & \underline{89.02}$\pm\scriptstyle{7.22}$ & \textbf{89.81}$\pm\scriptstyle{7.53}$ & 78.40$\pm\scriptstyle{22.53}$ & 84.53$\pm\scriptstyle{8.35}$ & 78.98$\pm\scriptstyle{18.21}$ & 85.79$\pm\scriptstyle{8.05}$ \\
                          & AUC             & 71.69$\pm\scriptstyle{12.57}$ & \underline{93.70}$\pm\scriptstyle{2.29}$ & 92.27$\pm\scriptstyle{3.01}$ & 93.43$\pm\scriptstyle{3.33}$ & 77.91$\pm\scriptstyle{8.84}$ & \textbf{96.47}$\pm\scriptstyle{0.71}$ & 77.79$\pm\scriptstyle{12.89}$ & 90.32$\pm\scriptstyle{1.00}$ \\ \midrule
\multirow{2}{*}{NMT}      & ACC             & 51.08$\pm\scriptstyle{0.50}$ & 71.79$\pm\scriptstyle{1.00}$ & \underline{72.50}$\pm\scriptstyle{1.11}$ & 70.50$\pm\scriptstyle{1.18}$ & 64.11$\pm\scriptstyle{1.14}$ & \textbf{73.29}$\pm\scriptstyle{3.35}$ & 69.96$\pm\scriptstyle{0.54}$ & 64.96$\pm\scriptstyle{0.00}$ \\
                          & AUC             & 73.35$\pm\scriptstyle{3.77}$ & \textbf{86.48}$\pm\scriptstyle{1.27}$ & \underline{86.23}$\pm\scriptstyle{0.46}$ & 85.77$\pm\scriptstyle{1.24}$ & 82.85$\pm\scriptstyle{1.04}$ & 85.27$\pm\scriptstyle{0.32}$ & 84.64$\pm\scriptstyle{0.50}$ & 77.98$\pm\scriptstyle{0.00}$ \\ \midrule
\multirow{2}{*}{SEE}      & ACC             & 68.29$\pm\scriptstyle{4.08}$ & \underline{70.55}$\pm\scriptstyle{1.21}$ & 68.35$\pm\scriptstyle{2.22}$ & 70.13$\pm\scriptstyle{0.86}$ & 64.80$\pm\scriptstyle{0.68}$ & 67.09$\pm\scriptstyle{1.33}$ & 68.00$\pm\scriptstyle{1.07}$ & \textbf{73.42}$\pm\scriptstyle{0.46}$ \\
                          & AUC             & \underline{78.92}$\pm\scriptstyle{0.65}$ & 77.72$\pm\scriptstyle{0.50}$ & 74.11$\pm\scriptstyle{2.62}$ & 76.64$\pm\scriptstyle{0.31}$ & 71.78$\pm\scriptstyle{0.77}$ & 75.70$\pm\scriptstyle{0.59}$ & 72.90$\pm\scriptstyle{1.35}$ & \textbf{80.72}$\pm\scriptstyle{0.20}$ \\ \midrule
\multirow{2}{*}{Ave}      & ACC             & 68.98 & 75.28 & 73.43 & \underline{75.78} & 68.06 & 74.64 & 71.15 & \textbf{77.40} \\
                          & AUC             & 76.46 & 84.12 & 79.11 & 82.82 & 74.34 & \textbf{85.95} & 74.52 & \underline{85.01} \\ \midrule
                          \multicolumn{2}{c}{\# top1}             & 0              & 1         & 0            & 1            & 0            & 12            & 0            & 10             \\
                          \multicolumn{2}{c}{Average ranking}             & 5.92              & 3.29         & 5.21            & 3.63            & 6.21            & 2.63            & 6.17            & 2.96             \\\bottomrule
\end{tabular}%
}
\end{table}

\newpage
\section{Linear Probing Classification}
\label{app_linear_probing}

\subsection{Experimental Settings}

In linear probing, we freeze the pretrained backbone and optimize only a lightweight linear head on top of the frozen representations. For language-model-based baselines \cite{neurolm,elmmil}, we train a linear language head; for encoder-style baselines \cite{brainbert,biot,labram,cbramod,csbrain}, we train a linear classification head.

\subsection{Results}

\begin{table}[htb]
\caption{\textbf{Linear probing results.}}
\resizebox{\textwidth}{!}{%
\begin{tabular}{@{}cccccccccc@{}}
\toprule
\textbf{Dataset}          & \textbf{Metric} & \textbf{BrainBERT} & \textbf{BIOT} & \textbf{LaBraM} & \textbf{CBraMod} & \textbf{NeuroLM} & \textbf{ELM-MIL} & \textbf{CSBrain} & \textbf{BrainCoT} \\ \midrule
\multirow{2}{*}{ISRUC}    & ACC             & 62.78$\pm\scriptstyle{16.03}$        & 57.14$\pm\scriptstyle{10.18}$   & 33.87$\pm\scriptstyle{12.26}$      & \underline{76.46}$\pm\scriptstyle{1.57}$       & 35.48$\pm\scriptstyle{1.48}$       & 65.04$\pm\scriptstyle{6.20}$       & 66.39$\pm\scriptstyle{7.98}$       & \textbf{86.08}$\pm\scriptstyle{1.39}$        \\
                          & AUC             & \underline{91.80}$\pm\scriptstyle{0.70}$         & 88.22$\pm\scriptstyle{5.59}$    & 76.09$\pm\scriptstyle{11.90}$      & 75.88$\pm\scriptstyle{10.13}$      & 83.79$\pm\scriptstyle{5.61}$       & 86.14$\pm\scriptstyle{5.77}$       & 59.83$\pm\scriptstyle{4.55}$       & \textbf{96.69}$\pm\scriptstyle{0.44}$        \\ \midrule
\multirow{2}{*}{Dreams}   & ACC             & 68.56$\pm\scriptstyle{4.15}$         & 71.59$\pm\scriptstyle{2.60}$    & 41.75$\pm\scriptstyle{2.59}$       & 70.24$\pm\scriptstyle{3.94}$       & 62.82$\pm\scriptstyle{2.42}$       & 40.93$\pm\scriptstyle{2.98}$       & \underline{73.11}$\pm\scriptstyle{3.33}$       & \textbf{80.69}$\pm\scriptstyle{2.14}$        \\
                          & AUC             & 88.43$\pm\scriptstyle{1.29}$         & 89.39$\pm\scriptstyle{1.30}$    & 49.66$\pm\scriptstyle{0.40}$       & 60.78$\pm\scriptstyle{2.58}$       & 82.78$\pm\scriptstyle{1.08}$       & \underline{90.40}$\pm\scriptstyle{1.47}$       & 50.34$\pm\scriptstyle{0.41}$       & \textbf{94.08}$\pm\scriptstyle{0.61}$        \\ \midrule
\multirow{2}{*}{Siena}    & ACC             & \textbf{52.79}$\pm\scriptstyle{14.12}$        & 50.95$\pm\scriptstyle{13.50}$   & 36.88$\pm\scriptstyle{19.42}$      & 50.68$\pm\scriptstyle{14.11}$      & 50.76$\pm\scriptstyle{10.96}$      & 49.87$\pm\scriptstyle{12.03}$      & \underline{52.08}$\pm\scriptstyle{14.84}$      & 49.43$\pm\scriptstyle{6.36}$        \\
                          & AUC             & 61.49$\pm\scriptstyle{7.35}$         & 60.83$\pm\scriptstyle{3.74}$    & 59.52$\pm\scriptstyle{4.43}$       & 59.33$\pm\scriptstyle{5.66}$       & \underline{63.14}$\pm\scriptstyle{5.35}$       & 56.29$\pm\scriptstyle{5.96}$       & 55.42$\pm\scriptstyle{1.89}$       & \textbf{65.09}$\pm\scriptstyle{5.60}$        \\ \midrule
\multirow{2}{*}{SanDiego} & ACC             & 64.38$\pm\scriptstyle{14.34}$        & 66.50$\pm\scriptstyle{1.67}$    & 67.62$\pm\scriptstyle{1.71}$       & \underline{67.95}$\pm\scriptstyle{3.80}$       & 61.59$\pm\scriptstyle{2.23}$       & 66.72$\pm\scriptstyle{2.64}$       & 66.43$\pm\scriptstyle{2.18}$       & \textbf{72.20}$\pm\scriptstyle{4.74}$        \\
                          & AUC             & 64.20$\pm\scriptstyle{11.78}$        & 53.76$\pm\scriptstyle{23.68}$   & 56.02$\pm\scriptstyle{5.78}$       & \underline{65.15}$\pm\scriptstyle{5.87}$       & 62.38$\pm\scriptstyle{4.71}$       & 60.42$\pm\scriptstyle{9.90}$       & 53.86$\pm\scriptstyle{2.90}$       & \textbf{73.77}$\pm\scriptstyle{4.12}$        \\ \midrule
\multirow{2}{*}{ADHD}     & ACC             & 67.07$\pm\scriptstyle{8.17}$         & 64.90$\pm\scriptstyle{3.53}$    & \textbf{77.98}$\pm\scriptstyle{5.85}$       & 64.46$\pm\scriptstyle{3.27}$       & 69.19$\pm\scriptstyle{6.15}$       & 66.26$\pm\scriptstyle{4.09}$       & 68.23$\pm\scriptstyle{2.57}$       & \underline{74.85}$\pm\scriptstyle{6.07}$        \\
                          & AUC             & 65.90$\pm\scriptstyle{5.66}$         & 71.42$\pm\scriptstyle{8.40}$    & 69.56$\pm\scriptstyle{5.35}$       & \textbf{83.14}$\pm\scriptstyle{6.66}$       & 69.01$\pm\scriptstyle{6.84}$       & 74.59$\pm\scriptstyle{4.15}$       & 71.63$\pm\scriptstyle{6.86}$       & \underline{79.69}$\pm\scriptstyle{5.70}$        \\ \midrule
\multirow{2}{*}{Schizo}   & ACC             & 73.86$\pm\scriptstyle{17.84}$        & 61.99$\pm\scriptstyle{5.69}$    & \underline{77.38}$\pm\scriptstyle{6.90}$       & 72.54$\pm\scriptstyle{6.73}$       & 54.89$\pm\scriptstyle{11.42}$      & 61.53$\pm\scriptstyle{5.14}$       & 62.97$\pm\scriptstyle{3.99}$       & \textbf{77.39}$\pm\scriptstyle{10.43}$       \\
                          & AUC             & 76.45$\pm\scriptstyle{9.98}$         & 79.04$\pm\scriptstyle{19.73}$   & 63.10$\pm\scriptstyle{7.80}$       & \textbf{83.46}$\pm\scriptstyle{8.15}$       & 61.65$\pm\scriptstyle{7.45}$       & 67.10$\pm\scriptstyle{6.53}$       & 53.57$\pm\scriptstyle{11.89}$      & \underline{83.41}$\pm\scriptstyle{9.65}$        \\ \midrule
\multirow{2}{*}{APAVA}    & ACC             & 73.01$\pm\scriptstyle{21.72}$        & 64.35$\pm\scriptstyle{5.65}$    & 70.59$\pm\scriptstyle{10.45}$      & \textbf{80.30}$\pm\scriptstyle{14.38}$      & 57.87$\pm\scriptstyle{3.78}$       & 64.97$\pm\scriptstyle{13.41}$      & 65.85$\pm\scriptstyle{8.70}$       & \underline{79.90}$\pm\scriptstyle{17.62}$       \\
                          & AUC             & \underline{92.60}$\pm\scriptstyle{4.42}$         & 78.46$\pm\scriptstyle{26.39}$   & 68.22$\pm\scriptstyle{12.12}$      & 79.26$\pm\scriptstyle{13.34}$      & 69.39$\pm\scriptstyle{17.69}$      & 76.69$\pm\scriptstyle{8.43}$       & 56.20$\pm\scriptstyle{3.74}$       & \textbf{94.39}$\pm\scriptstyle{4.98}$        \\ \midrule
\multirow{2}{*}{ADFSU}    & ACC             & 90.76$\pm\scriptstyle{3.93}$         & 87.98$\pm\scriptstyle{3.26}$    & \textbf{94.26}$\pm\scriptstyle{3.10}$       & \underline{93.68}$\pm\scriptstyle{3.29}$       & 83.41$\pm\scriptstyle{2.89}$       & 92.70$\pm\scriptstyle{2.93}$       & 87.02$\pm\scriptstyle{2.56}$       & 91.14$\pm\scriptstyle{5.79}$        \\
                          & AUC             & \textbf{94.06}$\pm\scriptstyle{2.53}$         & 90.20$\pm\scriptstyle{4.22}$    & 74.88$\pm\scriptstyle{7.21}$       & \underline{93.11}$\pm\scriptstyle{5.89}$       & 91.34$\pm\scriptstyle{6.15}$       & 50.83$\pm\scriptstyle{7.20}$       & 55.18$\pm\scriptstyle{13.15}$      & 91.67$\pm\scriptstyle{5.19}$        \\ \midrule
\multirow{2}{*}{IEDS}     & ACC             & \underline{69.63}$\pm\scriptstyle{9.52}$         & 66.77$\pm\scriptstyle{11.33}$   & 66.77$\pm\scriptstyle{12.24}$      & 66.40$\pm\scriptstyle{12.21}$      & 64.09$\pm\scriptstyle{8.04}$       & 66.47$\pm\scriptstyle{9.42}$       & 67.57$\pm\scriptstyle{11.54}$      & \textbf{69.70}$\pm\scriptstyle{7.81}$        \\
                          & AUC             & 50.12$\pm\scriptstyle{1.46}$         & \textbf{62.19}$\pm\scriptstyle{3.61}$    & 56.53$\pm\scriptstyle{2.16}$       & 57.98$\pm\scriptstyle{5.02}$       & 56.85$\pm\scriptstyle{2.38}$       & 58.68$\pm\scriptstyle{1.51}$       & 54.14$\pm\scriptstyle{1.54}$       & \underline{60.46}$\pm\scriptstyle{3.64}$        \\ \midrule
\multirow{2}{*}{Mayo}     & ACC             & 72.41$\pm\scriptstyle{27.52}$        & 72.81$\pm\scriptstyle{27.47}$   & 72.41$\pm\scriptstyle{27.52}$      & \underline{78.71}$\pm\scriptstyle{19.81}$      & 72.41$\pm\scriptstyle{27.52}$      & 75.32$\pm\scriptstyle{25.48}$      & 76.91$\pm\scriptstyle{22.55}$      & \textbf{86.22}$\pm\scriptstyle{9.56}$        \\
                          & AUC             & \underline{78.32}$\pm\scriptstyle{9.42}$         & 78.14$\pm\scriptstyle{17.25}$   & 64.71$\pm\scriptstyle{8.27}$       & 74.20$\pm\scriptstyle{11.01}$      & 69.21$\pm\scriptstyle{8.88}$       & 77.98$\pm\scriptstyle{4.50}$       & 55.14$\pm\scriptstyle{2.88}$       & \textbf{91.11}$\pm\scriptstyle{1.99}$        \\ \midrule
\multirow{2}{*}{NMT}      & ACC             & 59.43$\pm\scriptstyle{1.76}$         & 54.75$\pm\scriptstyle{1.40}$    & \textbf{67.86}$\pm\scriptstyle{0.41}$       & 54.36$\pm\scriptstyle{0.51}$       & 58.45$\pm\scriptstyle{1.65}$       & 58.76$\pm\scriptstyle{1.07}$       & 55.38$\pm\scriptstyle{0.26}$       & \underline{64.68}$\pm\scriptstyle{0.37}$        \\
                          & AUC             & \underline{78.69}$\pm\scriptstyle{0.07}$         & 73.95$\pm\scriptstyle{0.69}$    & 70.94$\pm\scriptstyle{1.27}$       & 70.45$\pm\scriptstyle{0.70}$       & 73.91$\pm\scriptstyle{0.14}$       & 75.64$\pm\scriptstyle{0.14}$       & 68.26$\pm\scriptstyle{1.01}$       & \textbf{81.04}$\pm\scriptstyle{0.73}$        \\ \midrule
\multirow{2}{*}{SEE}      & ACC             & 67.94$\pm\scriptstyle{2.35}$         & 65.55$\pm\scriptstyle{4.65}$    & 61.22$\pm\scriptstyle{3.69}$       & \underline{70.40}$\pm\scriptstyle{0.63}$       & 53.68$\pm\scriptstyle{2.42}$       & 63.78$\pm\scriptstyle{0.88}$       & 66.55$\pm\scriptstyle{1.06}$       & \textbf{72.69}$\pm\scriptstyle{0.28}$        \\
                          & AUC             & 76.55$\pm\scriptstyle{0.55}$         & \underline{78.24}$\pm\scriptstyle{2.60}$    & 70.91$\pm\scriptstyle{5.75}$       & 71.13$\pm\scriptstyle{1.18}$       & 68.44$\pm\scriptstyle{0.27}$       & 72.73$\pm\scriptstyle{1.70}$       & 54.36$\pm\scriptstyle{3.15}$       & \textbf{80.03}$\pm\scriptstyle{0.08}$        \\ \midrule
\multirow{2}{*}{Ave}      & ACC             & 68.55              & 65.44         & 64.05            & \underline{70.51}            & 60.39            & 64.36            & 67.37            & \textbf{75.42}             \\
                          & AUC             & \underline{76.55}              & 75.33         & 65.01            & 72.82            & 71.00            & 70.62            & 57.33            & \textbf{82.62}             \\  \midrule
                          \multicolumn{2}{c}{\# top1}             & 2              & 1         & 3            & 3            & 0            & 0            & 0            & 15             \\
                          \multicolumn{2}{c}{Average ranking}             & 3.79              & 4.56         & 5.35            & 4.00            & 5.96            & 4.88            & 5.75            & 1.71             \\ \bottomrule
\end{tabular}%
}
\end{table}

\newpage
\section{Few-shot Classification}
\label{app_few_shot}
\subsection{Experimental Settings}

In clinical brain-signal analysis, labeled data are often scarce because expert annotation is expensive, time-consuming, and requires domain knowledge. Therefore, the ability to adapt a pretrained model with only a few labeled examples is important for practical deployment in data-limited medical scenarios. To evaluate this capability, we compare few-shot classification with standard linear probing. Both settings follow the linear-probing paradigm, where the pretrained backbone is frozen and only the final prediction head is optimized. For few-shot classification, we construct a support set by randomly sampling $K\in\{1,2,4,8\}$ labeled examples per class from the training split, and train the same head on this support set. For linear probing, we train the head using the full training split.

\subsection{Results}
\begin{figure*}[htpb]
\begin{center}
\centerline{\includegraphics[width=\textwidth]{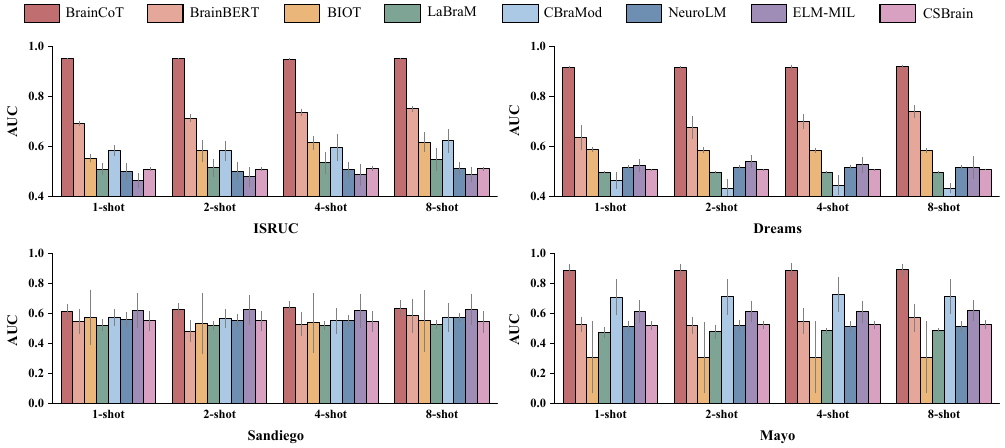}}
\caption{\textbf{Few-shot classification performance on four representative EEG/iEEG datasets}. We report AUC under 1/2/4/8-shot settings on ISRUC and Dreams (sleep stage classification), SanDiego (Parkinson’s disease detection), and Mayo (interictal epileptiform activity detection). Bars show mean performance and error bars indicate variability across runs/folds.}
\label{fig3_fewshot}
\end{center}
\vspace{-0.5cm}
\end{figure*}

\newpage
\section{t-SNE Visualization of Learned Representations}
\begin{figure*}[htpb]
    \centering
    \includegraphics[width=0.85\linewidth]{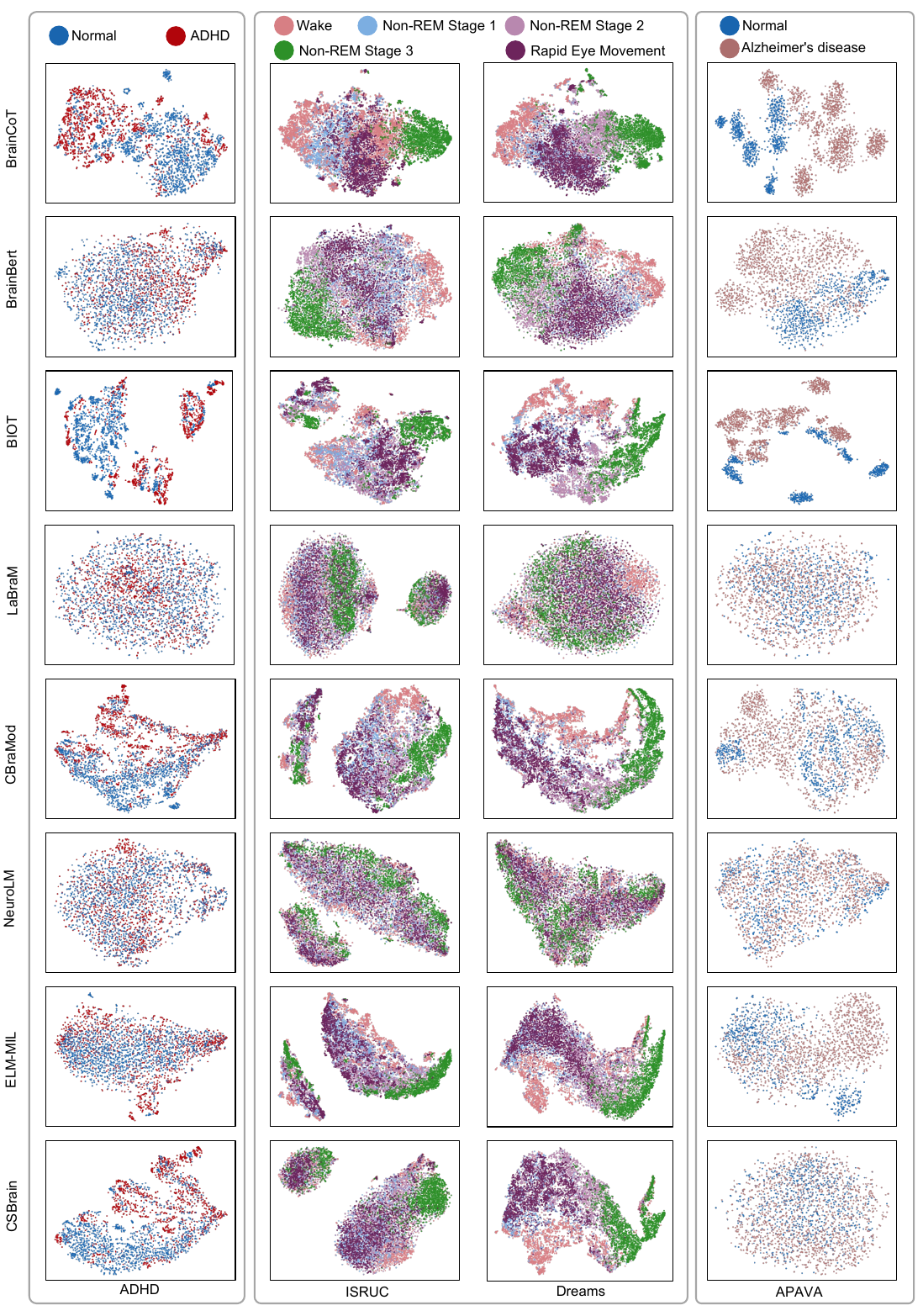}
    \caption{\textbf{t-SNE visualizations of learned representations across different datasets and models.}
    Each row corresponds to a backbone model, and each column corresponds to a dataset (ADHD-121, ISRUC, Dreams, and APAVA). Points are colored by ground-truth class labels.}
    \label{fig:tsne_baselines}
\end{figure*}

For a fair comparison, we freeze the backbone parameters for BrainCoT and all baselines, and extract their per-segment embeddings for visualization. We then apply t-SNE to project the embeddings into 2D, which provides a qualitative view of representation geometry and class separability without confounding effects from downstream fine-tuning. As shown in Fig.~\ref{fig:tsne_baselines} on four representative downstream datasets (ADHD, ISRUC, Dreams, and APAVA), BrainCoT tends to form more compact and better separated clusters across classes, while several baselines exhibit heavier overlap. This suggests that BrainCoT learns more discriminative and transferable representations, consistent with its stronger performance under low-adaptation or zero-adaptation settings.

\newpage
\section{Discussion}
\subsection{Limitations}
To our knowledge, BrainCoT is the first brain signal foundation model that supports \textbf{multi-task zero-shot brain signal analysis via evidence-anchored Chain-of-Thought reasoning}. While this work establishes a practical and verifiable reasoning paradigm, several limitations remain.
(1) \textbf{Efficiency and footprint.} As a native signal--language model, BrainCoT is not optimized for minimal resource usage, and its decoding-style pipeline can be less favorable in latency and memory compared with lightweight task-specific models.
(2) \textbf{Upper-bound performance under full supervision.} When full fine-tuning is allowed, BrainCoT does not consistently surpass the strongest end-to-end specialists on every dataset. Its primary advantage lies in plug-and-play generalization and reduced adaptation cost, rather than solely maximizing supervised ceilings.
(3) \textbf{Dependence on labeled corpora for pretraining.} Stage-\RNum{1} benefits from large-scale labeled datasets to construct instruction--answer pairs, whereas purely self-supervised pipelines can exploit unlabeled recordings more directly. We view this as a pragmatic trade-off for obtaining reliable zero-shot task execution and verifiable reasoning trajectories.

\subsection{Broader Impacts}

BrainCoT may improve the usability and transparency of brain-signal analysis by enabling multi-task zero-shot inference and neurometric-anchored reasoning. By reducing the need for target-data-specific annotation and adaptation, it may support more accessible EEG/iEEG analysis in research and data-limited settings, while its reasoning trajectories expose intermediate evidence cues rather than only final predictions.

However, BrainCoT should be used with caution and should not be treated as a standalone clinical diagnostic system. Brain signals differ across subjects, devices, centers, and clinical populations, so model errors may lead to unreliable conclusions without expert review.

Future use should also consider privacy and bias. Brain signals may contain sensitive health-related information, and the model may inherit biases from imbalanced datasets, recording protocols, or annotation practices. Therefore, any real-world use should require independent clinical validation, uncertainty reporting, fairness checks, and human expert oversight. BrainCoT is intended for research use and decision support, not autonomous diagnosis.

\subsection{Outlook}
BrainCoT points to a broader direction, moving from predicting labels to performing brain signal analysis with verifiable evidence. Looking ahead, we see three promising avenues.
(1) \textbf{Scaling and broader coverage.} Expanding to more diverse populations, acquisition hardware, and rare cohorts, ideally in multi-center prospective settings, will further validate robustness and may unlock new capabilities in open-world brain signal analysis.
(2) \textbf{Efficiency-oriented deployment.} Knowledge distillation \cite{knowledge_distillation}, lightweight backbones, and expert sparsification can translate evidence-anchored reasoning into real-time and edge-friendly systems, enabling practical use in wearable and bedside scenarios without sacrificing transparency.
(3) \textbf{Better use of unlabeled data.} Integrating self-supervised objectives with our instruction-and-reasoning framework, for example by coupling representation learning with evidence selection and trajectory supervision, could reduce reliance on labeled corpora while retaining verifiability. Together, these directions may turn evidence-anchored reasoning from a model feature into a general interface for trustworthy and scalable brain signal foundation models.

\clearpage

\end{document}